\documentclass[sn-mathphys,Numbered]{sn-jnl}

\usepackage{graphicx}%
\usepackage{multirow}%
\usepackage{amsmath,amssymb,amsfonts}%
\usepackage{amsthm}%
\usepackage{mathrsfs}%
\usepackage[title]{appendix}%
\usepackage{xcolor}%
\usepackage{tabularx}%
\usepackage{textcomp}%
\usepackage{manyfoot}%
\usepackage{booktabs}%
\usepackage{algorithm}%
\usepackage{algorithmicx}%
\usepackage{algpseudocode}%
\usepackage{listings}%

\usepackage{enumitem}
\usepackage{xr}

\theoremstyle{thmstyleone}%
\theoremstyle{thmstyletwo}%

\theoremstyle{thmstylethree}%
\newtheorem{definition}{Definition}%

\begin{document}

\title[Article Title]{DeepLINK-T: deep learning inference for time series data using knockoffs and LSTM}


\author[1]{\fnm{Wenxuan} \sur{Zuo}}\email{wzuo@usc.edu}

\author[1]{\fnm{Zifan} \sur{Zhu}}\email{zifanzhu@usc.edu}

\author[1]{\fnm{Yuxuan} \sur{Du}}\email{yuxuandu@usc.edu}

\author[3,4]{\fnm{Yi-Chun} \sur{Yeh}}\email{yichunyeh@alumni.usc.edu}

\author[3]{\fnm{Jed A.} \sur{Fuhrman}}\email{fuhrman@usc.edu}

\author[2]{\fnm{Jinchi} \sur{Lv}}\email{jinchilv@marshall.usc.edu}

\author*[2]{\fnm{Yingying} \sur{Fan}}\email{fanyingy@marshall.usc.edu}

\author*[1]{\fnm{Fengzhu} \sur{Sun}}\email{fsun@usc.edu}

\affil[1]{\orgdiv{Quantitative and Computational Biology Department}, \orgname{University of Southern California}, \orgaddress{\city{Los Angeles}, \postcode{90089}, \state{CA}, \country{USA}}}

\affil[2]{\orgdiv{Data Sciences and Operations Department}, \orgname{University of Southern California}, \orgaddress{\city{Los Angeles}, \postcode{90089}, \state{CA}, \country{USA}}}

\affil[3]{\orgdiv{Department of Biological Sciences}, \orgname{University of Southern California}, \orgaddress{\city{Los Angeles}, \postcode{90089}, \state{CA}, \country{USA}}}

\affil[4]{\orgdiv{Department of Global Ecology, Carnegie Institution for Science}, \orgname{Stanford University}, \orgaddress{\city{Stanford}, \postcode{94305}, \state{CA}, \country{USA}}}

\date{\today}


\abstract{
High-dimensional longitudinal time series data is prevalent across various real-world applications. Many such applications can be modeled as regression problems with high-dimensional time series covariates. Deep learning has been a popular and powerful tool for fitting these regression models. Yet, the development of interpretable and reproducible deep-learning models is challenging and remains underexplored. This study introduces a novel method, Deep Learning Inference using Knockoffs for Time series data (DeepLINK-T), focusing on the selection of significant time series variables in regression while controlling the false discovery rate (FDR) at a predetermined level. DeepLINK-T combines deep learning with knockoff inference to control FDR in feature selection for time series models, accommodating a wide variety of feature distributions. It addresses dependencies across time and features by leveraging a time-varying latent factor structure in time series covariates. Three key ingredients for DeepLINK-T are 1) a Long Short-Term Memory (LSTM) autoencoder for generating time series knockoff variables, 2) an LSTM prediction network using both original and knockoff variables, and 3) the application of the knockoffs framework for variable selection with FDR control. Extensive simulation studies have been conducted to evaluate DeepLINK-T's performance, showing its capability to control FDR effectively while demonstrating superior feature selection power for high-dimensional longitudinal time series data compared to its non-time series counterpart. DeepLINK-T is further applied to three metagenomic data sets, validating its practical utility and effectiveness, and underscoring its potential in real-world applications.
}

\keywords{Deep learning inference, high dimensionality, time series, knockoffs, autoencoder, LSTM, FDR control, power, metagenomics, DeepLINK}



\maketitle

\section{Introduction} \label{sec1}

Over the past decades, feature selection has become one of predominant techniques in various fields including bioinformatics \cite{saeys2007review, wang2016feature, love2014moderated, lin2020analysis}, economics \cite{zhang2014causal, zhao2016distributed}, and healthcare \cite{remeseiro2019review, nagarajan2021feature, rado2019performance}.  Its main goal is to search over a large set of potential variables, often consisting of both relevant and irrelevant ones, to pinpoint those that are crucial for the outcomes of interest. By selecting the most relevant covariates, it not only aids in simplifying models for easier interpretation but also plays a crucial role in reducing the complexity of data and improving computational efficiency \citep{jovic2015review} and reproducibility. This process is fundamental across a variety of machine learning applications, including but not limited to, classification, regression, and clustering tasks.


The prevalence of complex data types nowadays has spurred the development of advanced data science tools, such as random forests and deep neural networks, which are among the most widely used for analyzing modern big data. However, the complexity of these methods often produces data analysis results that are challenging to interpret and replicate. Integrating feature selection into the model development process can significantly aid in simplifying these models, and hence alleviate the concerns of interpretability and reproducibility. Much effort has been devoted to tackling the intricate challenge of feature selection. For example, the random forests (RF) approach \cite{breiman2001random} utilizes information gain for feature selection, but it may be susceptible to a high false discovery rate (FDR) in the setting of high-dimension inference; see, e.g., \cite{2022HRF} for some recent theoretical developments on high-dimensional random forests. Seminal works based on p-values \cite{benjamini1995controlling, benjamini2001control} and various extensions \cite{ ignatiadis2016data, scott2015false} have been introduced to access the feature importance. However, the efficacy of p-value-based methods is restricted to simpler model settings where obtaining p-values is plausible. For intricate models such as deep neural networks, it is highly difficult or even completely impossible to calculate p-values, thereby posing a formidable challenge for feature selection using p-value-based methods in such contexts. Moreover, most existing works focus mainly on the independent data setting and are not able to accommodate the serial dependence in time series data.


Recently, a novel approach named the knockoff filter was introduced in \cite{barber2015controlling}, designed for FDR control in low-dimensional Gaussian linear models without relying on p-values. Subsequently, the model-X knockoffs framework \cite{2018MXK} was proposed to generalize the knockoff filter to general high-dimensional nonlinear models with arbitrary dependency structure of the predictors. The model-X knockoffs framework essentially acts as a wrapper, seamlessly integrating with any feature selection methods that yield feature importance measures satisfying specific conditions. Recently, Lu et al. \cite{2018DeepPINK} proposed DeepPINK to combine the model-X knockoffs and deep neural networks for features characterized by joint Gaussian distributions, and Zhu et al. \cite{2021DeepLINK} introduced DeepLINK to generalize the distribution of input features via a latent factor model strategy for non-time series data. More recent developments on the extensions of model-X knockoffs inference include \cite{2020RANK,2020IPAD,2021KIMI,2024TSKI,2024ARK}. In particular, our current work was motivated by the idea of utilizing the latent factor structure in the recent work of IPAD knockoffs in \cite{2020IPAD}.

These existing knockoffs based methods have primarily concentrated on scenarios involving independent observations, thus failing to address the widespread phenomenon of serial dependence that is commonly encountered in data from various domains, including ecology, medicine, and finance. Incorporating temporal information into the feature selection process has the potential to produce more accurate and interpretable variable selection results. Motivated from this, we introduce a novel method for deep learning inference using knockoffs specifically tailored for time series data, named DeepLINK-T.  There are three key components in DeepLINK-T: 1) a Long Short-Term Memory (LSTM) autoencoder for generating time series knockoff variables incorporating the serial dependence, 2) an LSTM prediction network for fitting a regression model using augmented features formed by both original and knockoff features, and 3) the application of the knockoffs framework for variable selection with FDR control. The use of LSTM networks enables DeepLINK-T to model and leverage the serial dependence in data, and hence produces more accurate variable selection outcomes. This is verified through intensive simulation studies in our paper.  Additionally, we examine the robustness of DeepLINK-T to some misspecification of the time series latent factor model; our findings affirm that DeepLINK-T maintains control over FDR while retaining high variable selection power, provided that the number of training epochs is sufficiently large. Finally, the practical usage of DeepLINK-T is tested on three metagenomic data sets. Specifically, DeepLINK-T pinpointed \textit{Parabacteroides} as a key genus that alters the abundance level of \textit{Bacteroides} in the gastrointestinal tract of early infants. Furthermore, our observations revealed that \textit{Pyramimonas} and \textit{Heterosigma} were notably associated with the concentration of chlorophyll-a in a marine metagenomic time series. Additionally, \textit{Ruminiclostridum} and \textit{Rothia} were identified to be significantly correlated with two distinct types of dietary glycans treatment. Overall, we anticipate that DeepLINK-T will contribute to the feature selection analysis of time series data, providing novel insights into unexplored aspects of the microbiome.   

\section{Method} \label{sec:methods}

\subsection{Model setting} \label{new.sec2.1}

Consider the problem of high-dimensional supervised learning, which involves analyzing longitudinal time series observations on $m$ subjects. 
The time series data for each subject is represented as $(\mathbf{X}_i, \mathbf{y}_i), i=1,\ldots, m$, where $\mathbf{X}_i=(\mathbf{x}_{i1},\ldots, \mathbf{x}_{in})^T \in \mathbb{R}^{n\times p}$ is the feature matrix containing $n$ time points and $p$ features for the $i$th subject, and the response vector over the $n$ time points is represented by $\mathbf{y}_i = (y_{i1},\ldots,y_{in})^T \in \mathbb{R}^{n}$. To simplify the technical presentation, let us assume that $(\mathbf{X}_i, \mathbf{y}_i)$'s are independent and identically distributed (i.i.d.) across $i \in \{1, \ldots, m\}$. The full set of features is represented as $\{1, \ldots, p\}$. We assume that given each row of $\mathbf{X}_i$, the corresponding conditional distribution of response $y_{ik}, k=1,\ldots,n$, depends only on a common small subset of features, and our objective is to identify such Markov blanket, which is the smallest subset $S_0$ of $\{1, \ldots, p\}$ that makes $y_{ik}$ independent of all other features conditional on those in $S_0$, while controlling the feature selection error rate under a predefined level. 

Precisely, we aim to select important features in $S_0$ while keeping the false discovery rate (FDR) \cite{benjamini1995controlling} controlled, where the FDR is defined as
\begin{equation} \label{eq:fdr}
\text{FDR}=\mathbb{E}[\text{FDP}]\ \text{ with } \text{FDP}=\frac{|\hat{S}\cap S_{0}^{c}|}{\text{max}\{|\hat{S}|,1\}}.
\end{equation}
Here, $|\cdot|$ represents the cardinality of a set, FDP is the false discovery proportion, and $\hat{S}$ stands for the set of features
selected by a learning procedure. A slightly modified version of FDR (mFDR) is defined as 
\begin{equation} \label{eq:mfdr}
\text{mFDR}=\mathbb{E}
\left[
\frac{|\hat{S}\cap S_{0}^{c}|}{|\hat{S}|+q^{-1}}
\right].
\end{equation}
Here, $q\in (0,1)$ is the targeted FDR level. It is evident that FDR is more conservative than mFDR since the mFDR is always under control if the FDR is under control. Additionally, we utilize another important metric, the power, to measure the ability of the learning procedure in discovering the true features. Power is defined as the expectation of the true discovery proportion (TDP)
\begin{equation} \label{eq:power}
\text{Power}=\mathbb{E}[\text{TDP}] \ \text{ with } \text{TDP}=\frac{|\hat{S}\cap S_{0}|}{|S_0|}.
\end{equation}

As mentioned in the Introduction, the p-value based FDR approaches are not directly applicable to deep learning models due to the lack of valid p-value calculation methods caused by the complicated model structure.  Existing knockoffs methods have focused mainly on the independent data setting. The recent development of DeepLINK \cite{2021DeepLINK} has addressed deep learning inference for independent data without serial dependency. In this paper, we propose to combine the model-X knockoffs framework \cite{2018MXK} and the long short-term memory (LSTM) network \cite{hochreiter1997long} to achieve the FDR control in terms of feature selection for longitudinal time series applications.

To make our idea concrete, let us assume that response $y_{it}$ for the $i$th subject at the $t$th time point depends on $p$-dimensional feature vector $\mathbf{x}_{it}$ through the following regression model 
\begin{equation} \label{eq:regression_model}
    y_{it} = l(\mathbf{x}_{it}) + \varepsilon_{it},\ i = 1, \cdots, m,\ t = 1, \cdots, n,
\end{equation}
where $l$ stands for some unknown true regression function that can be either linear or nonlinear, and $\varepsilon_{it}$'s are 
i.i.d. model errors. To accommodate the potentially strong feature dependency, we focus on the scenario when the high-dimensional feature vector $\mathbf{x}_{it}, i=1,\ldots, m$, at time point $t$ depends on some low-dimensional latent factor vector $\mathbf{f}_{it}\in \mathbb{R}^r$ in a potentially nonlinear manner. Here, we assume that $r \ll p$. For clear presentation, let us first consider the case of one time series (i.e., $m = 1$) and use $\mathbf{x}_{t}$ and $\mathbf{f}_{t}$ to denote $\mathbf{x}_{it}$ and $\mathbf{f}_{it}$, respectively. We consider a factor structure for feature vector $\mathbf{x}_{t}$  described by equation
\begin{equation} \label{eq:factor_structure}
\mathbf{x}_{t}=g(\mathbf{f}_{t})+\boldsymbol\epsilon_{t}, \ t=1,\ldots,n.
\end{equation}
Here, $g$ represents some unknown link function whose coordinates can take either linear or nonlinear functional forms, and $\boldsymbol\epsilon_{t}=(\epsilon_{t,1},\cdots,\epsilon_{t,p})^T$ is the error vector  from some distribution $f_{\theta}$ and is independent of $\mathbf f_t$ and $\boldsymbol\epsilon_{t'}$ at other time points $t'\neq t$. Here, the distribution $f_{\theta}$ is assumed to be parametric with some unknown parameter (vector) $\theta$.
Considering the potential time dependency among the latent factors, we further assume the following dependency structure for the latent factors 
\begin{equation} \label{eq:factor_weighted}
\begin{aligned}
&\mathbf{f}_{1} = \mathbf{f}^{raw}_{1}, \\
&\mathbf{f}_{t} = w_0\mathbf{f}^{raw}_{t - 1} + w_1\mathbf{f}^{raw}_{t}, \ t = 2, \cdots, n,
\end{aligned}
\end{equation}
where $\mathbf{f}^{raw}_{t}$'s are vectors of the raw latent factors, and $w_0, w_1$ are the weights for the time dependency satisfying $w_0+w_1=1$. Here, we assume that $\mathbf{f}^{raw}_{t}$'s are drawn from a certain distribution and details are specified in the simulation designs. To simplify the notation, let $\mathbf{X}=(\mathbf{x}_{1}, \ldots, \mathbf{x}_{n})^T$ be the $n\times p$ design matrix for a given subject, $\mathbf{C}=(g(\mathbf{f}_{1}), \ldots, g(\mathbf{f}_{n}))^T$ the $n\times p$ factor matrix, and $\mathbf{E}=(\boldsymbol\epsilon_{1}, \ldots, \boldsymbol\epsilon_{n})^T$ the error matrix for the factor model. Then the factor model in Equation \eqref{eq:factor_structure} becomes
\begin{equation} \label{eq:factor_structure_matrix}
    \mathbf{X}=\mathbf{C}+\mathbf{E}.
\end{equation}

Hence, our objective is to develop a new deep learning  feature selection method that keeps FDR controlled at the target level $q$ under the time series model setting outlined in Equations \eqref{eq:regression_model}--\eqref{eq:factor_structure_matrix}.

\subsection{The model-X knockoffs framework} \label{new.sec2.2}

The idea of knockoffs inference was initially proposed in \cite{barber2015controlling} for the case of low-dimensional Gaussian linear models. Then the model-X knockoffs framework \cite{2018MXK} generalizes this approach to general high-dimensional nonlinear models and achieves theoretically guaranteed FDR control in arbitrary
dimensions and for arbitrary dependence structure of response on the features. Concisely, the model-X knockoffs framework constructs the knockoff variables that imitate the dependence structure of the original variables, but are conditionally independent of the response given the original features. These knockoff variables function as control variables, aiding in the assessment of the original variables' importance. The FDR control is achieved through the construction of what are referred to as the knockoff statistics -- a concept that will be explained subsequently -- which serve as measures of the original variables' importance. 

\begin{definition}[\cite{2018MXK}]\label{def:knockoffs}
Consider covariate vector $\mathbf{x}=(x_1,\ldots,x_p)^T$ and response $y$. 
The model-X knockoff variables for variables in $\mathbf{x}$ are random variables in $\tilde{\mathbf{x}}=(\tilde{x}_1,\ldots,\tilde{x}_p)^T$ that satisfy the following two properties: 

\begin{enumerate}[label=\arabic*), leftmargin=2\parindent] 
\item For any subset $S \subset \{1, 2,\ldots, p\}, (\mathbf{x}, \tilde{\mathbf{x}})_{\text{swap}(S)}\stackrel{d}{=}(\mathbf{x}, \tilde{\mathbf{x}})$, where $\text{swap}(S)$ is the operator that swaps the original variable $x_j$ and its knockoff counterpart $\tilde{x}_j$ for each $j \in S$, and $\stackrel{d}{=}$ stands for equal in distribution.
\item The knockoff variables $\tilde{\mathbf{x}}$ is independent of response $y$ given the original variables $\mathbf{x}$.
\end{enumerate}
\end{definition}

Let $\tilde{\mathbf{X}} = (\tilde{\mathbf{x}}_{1},\ldots, \tilde{\mathbf{x}}_{n})^T$ be the constructed knockoffs data matrix. The model-X knockoffs framework constructs knockoff statistics $W_1, \ldots, W_p$ to evaluate the importance of the original variables. For each $j \in \{1, \ldots, p\}$, $W_j=w_j([\mathbf{X}, \tilde{\mathbf{X}}], \mathbf{y})$ is a function of the augmented data matrix $[\mathbf{X}, \tilde{\mathbf{X}}]$ and response $\mathbf{y}$ satisfying the \textit{sign-flip property}
\begin{equation} \label{eq:sign_flip}
w_j([\mathbf{X}, \tilde{\mathbf{X}}]_{\text{swap}(S)}, \mathbf{y})=
\begin{cases}
w_j([\mathbf{X}, \tilde{\mathbf{X}}],\mathbf{y}), &j \notin S,\\
-w_j([\mathbf{X}, \tilde{\mathbf{X}}],\mathbf{y}), &j \in S,
\end{cases}
\end{equation}
where $S \subset \{1, \ldots, p\}$ is any given subset. Intuitively, the ideal knockoff statistics quantify the significance of original variables, with high positive values indicating their importance. On the other hand, for unimportant features not in $S_0$, the corresponding $W_j$'s are expected to have small magnitude and be symmetrically distributed around zero. 

Finally, the model-X knockoffs framework selects important features as $\hat{S}=\{j:W_j\geq t\}$ with $t$ being either the knockoff or knockoff+ threshold \cite{2018MXK} defined respectively as
\begin{equation} \label{eq:threshold}
    T=\min \left\{ t>0:\frac{|\{j:W_j\leq -t\}|}{\max\{|\{j:W_j\geq t\}|,1\}} \leq q \right\},
\end{equation}
\begin{equation} \label{eq:threshold+}
    T_+=\min \left\{ t>0:\frac{1+|\{j:W_j\leq -t\}|}{\max\{|\{j:W_j\geq t\}|,1\}} \leq q \right\}.
\end{equation}

Here, no features will be selected if $T=\infty$ or $T_+=\infty$. It has been proved in \cite{2018MXK} that under some regularity conditions, the model-X knockoffs framework controls the FDR in finite samples with the knockoff+ threshold, and controls the mFDR with the knockoff threshold.

It is important to note that the existing developments reviewed above are for independent data without serial correlation. Some theoretical development for time series knockoffs inference was recently established in \cite{2024TSKI} when $m=1$, under some stationarity assumption of time series data. Yet, the results in \cite{2024TSKI} are intended for laying down some general theoretical and methodological foundation and are not specific for the complex data and model settings considered in our paper.       

\subsection{DeepLINK-T: a new deep learning inference method for time series data} \label{new.sec2.3}

The preceding sections have highlighted the crucial elements of the model-X knockoffs framework: the construction of knockoff variables and the construction of knockoff statistics. In this subsection, we introduce DeepLINK-T, a deep learning-based statistical inference framework using knockoffs and LSTM for longitudinal time series data, for achieving the knockoffs construction task and feature selection task with controlled FDR. Briefly, the DeepLINK-T framework contains two components: 1) an LSTM autoencoder network for generating knockoff variables and 2) another LSTM prediction network for the knockoff statistics construction.

Our motivation comes from the recent works of IPAD \cite{2020IPAD} and DeepLINK \cite{2021DeepLINK} on knockoffs inference. Equation \eqref{eq:factor_structure_matrix} clearly shows that the knockoff variables can be directly constructed satisfying the two properties in Definition \ref{def:knockoffs} if we have the realization $\mathbf{C}$ and the distribution $f_\theta$ of the error matrix $\mathbf{E}$; we can generate an independent copy of the error matrix $\mathbf E$ from $f_\theta$ and then add it back to the realization $\mathbf C$ to form the knockoff variable matrix $\tilde{\mathbf X}$. However, in practice, it is challenging to obtain the realization $\mathbf{C}$, and we need to estimate $\mathbf{C}$ via a particular procedure. It is crucial to note that even though $\mathbf{C}$ is a random matrix, the construction of valid knockoff variables requires us to know the specific realization $\mathbf{C}$ rather than the distribution of realization $\mathbf{C}$. This is because if $\hat{\mathbf{C}}$ is independently drawn from the distribution of $\mathbf C$, the first property in the definition of model-X knockoff variables could be violated due to the independence of $\hat{\mathbf{C}}$ and $\mathbf{C}$.

\begin{figure}[t]%
\centering
\includegraphics[width=0.9\textwidth]{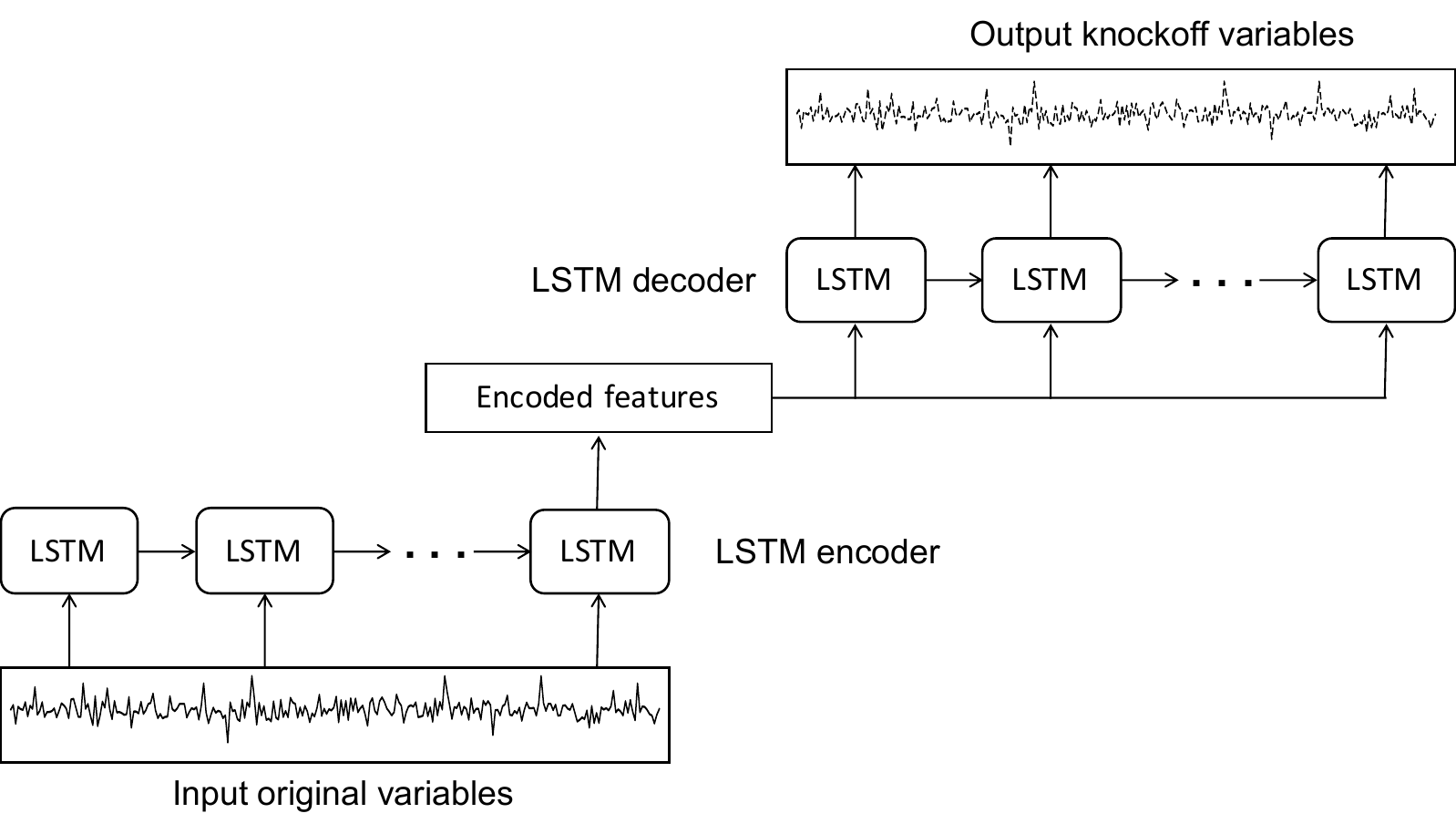}
\caption{The structure of the LSTM autoencoder.}\label{fig:LSTM_autoencoder}
\end{figure}

To overcome this challenge, we propose to use an LSTM autoencoder \cite{sagheer2019unsupervised} for generating the knockoff variables. The structure of the LSTM autoencoder is shown in Figure \ref{fig:LSTM_autoencoder}. Take one subject (time series) with $m=1$ as an example. We train an LSTM autoencoder with design matrix $\mathbf{X}$ as both the input and the target of the neural network. Let $\hat{\mathbf{C}}$ be the output matrix of the LSTM autoencoder for reconstruction. The estimated knockoff variables $\hat{\mathbf{X}}$ can be constructed by
\begin{equation} \label{eq:knockoff_variable_est}
\tilde{\mathbf{X}}=\hat{\mathbf{C}}+\tilde{\mathbf{E}}.
\end{equation}
Here, $\tilde{\mathbf{E}}$ is the regenerated error matrix with rows independently drawn from the estimated distribution $f_{\hat{\theta}}$ of $\boldsymbol\epsilon_{t}$. For instance, if the entries of the error matrix $\mathbf E$ are i.i.d. and follow the normal distribution with mean $0$ and variance $\sigma^2$, we can estimate $\theta=\sigma^2$ as
\begin{equation} \label{eq:normal_variance_est}
    \hat{\theta}=\frac{1}{np}\sum_{t=1}^{n}\sum_{k=1}^{p}\hat{e}_{tk}^2,
\end{equation}
where $\hat{e}_{tk}$'s are the entries of the residual matrix $\hat{\mathbf{E}}=\mathbf{X}-\hat{\mathbf{C}}$. 

In the existence of multiple subjects (i.e., $m>1$), each subject is treated as an input batch for the LSTM autoencoder. Stateless LSTM layers are applied to ensure that the initial state of the current batch will not be related to the last state of the previous batch. The variance estimator $\hat{\theta}_i$ is calculated separately for each batch $i=1,\cdots, m$, and then the error matrix $\tilde{\mathbf{E}}_i$ for each batch is independently drawn from its  estimated marginal distribution $f_{\hat{\theta}_i}$. Since different subjects are treated separately in generating the knockoff variables, we will continue introducing our method assuming $m=1$ for simplicity of notation whenever there is no confusion. 

\begin{figure}[t]%
\centering
\includegraphics[width=0.9\textwidth]{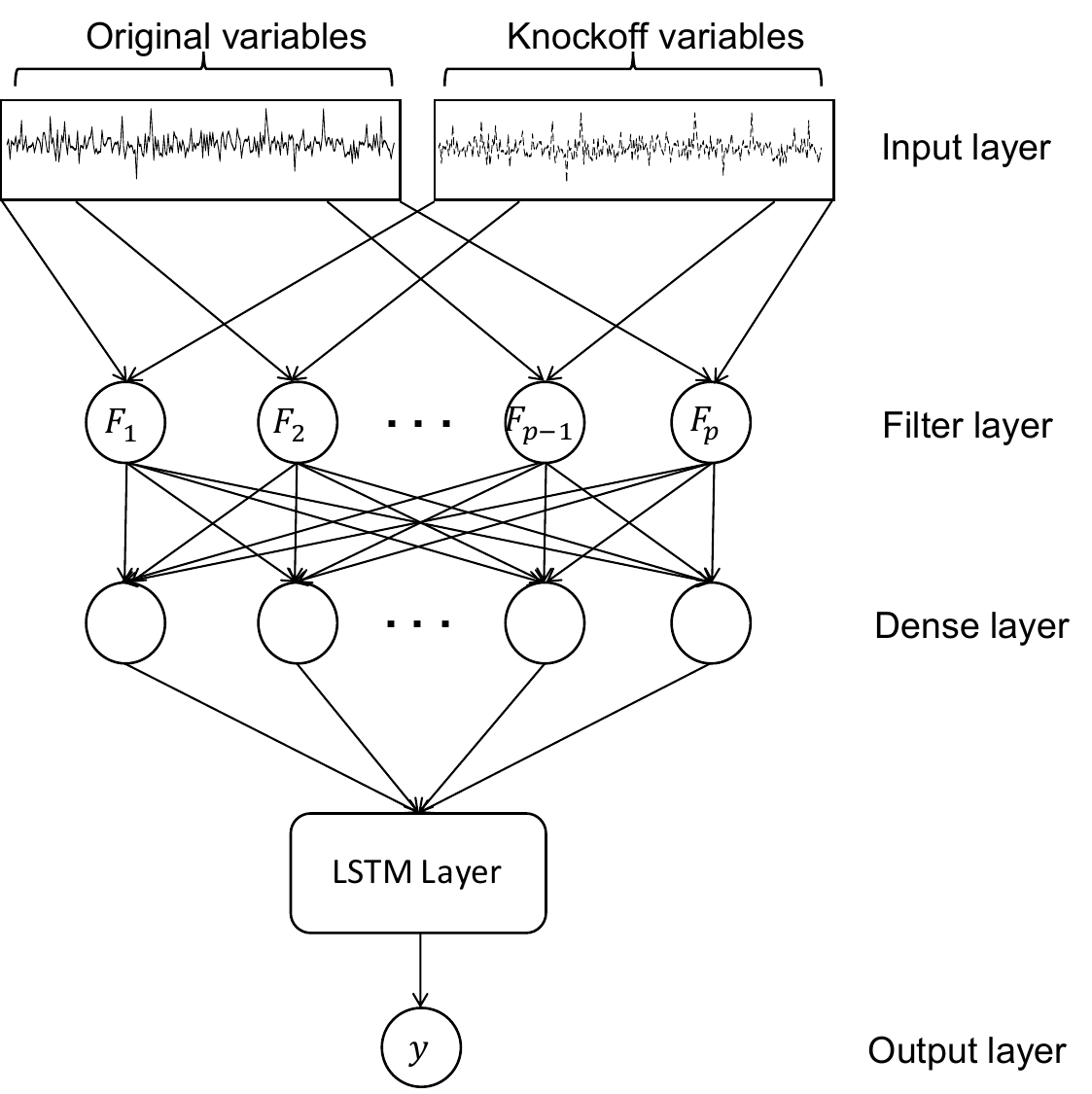}
\caption{The structure of the LSTM prediction network.}\label{fig:LSTM_prediction_nn}
\end{figure}

As discussed in the previous section, the ideal knockoff statistics are able to accurately measure the importance of original features. Additionally, the relationship between response variable $y_{it}$ and features $\mathbf{x}_{it}$ in Equation \eqref{eq:regression_model} could be either linear or nonlinear. To achieve these objectives, we propose to utilize the weights from a learned LSTM prediction network to construct the knockoff statistics. The structure of the LSTM prediction network is shown in Figure \ref{fig:LSTM_prediction_nn}. 

Specifically, after generating the knockoff variable matrix $\tilde{\mathbf{X}}$ from the LSTM autoencoder, we form the augmented data matrix $[\mathbf{X}, \tilde{\mathbf{X}}]$. To integrate the information from the original variables $\mathbf{X}$ and the knockoff variables $\tilde{\mathbf{X}}$, we adopt the idea of DeepPINK \cite{2018DeepPINK} to form an LSTM prediction network (Figure \ref{fig:LSTM_prediction_nn}).  In Figure \ref{fig:LSTM_prediction_nn}, each of the original features and their knockoff counterparts were paired together using a plugin pairwise-coupling layer to ensure fair competition in their importance for predicting the response variable, and this layer is named as the filter layer for the ease of reference. The output from the filter layer is then fed to a dense layer followed by an LSTM layer with the output being predicted response vector $\mathbf y_{i}$. 

\begin{figure}[t]%
\centering
\includegraphics[width=0.8\textwidth]{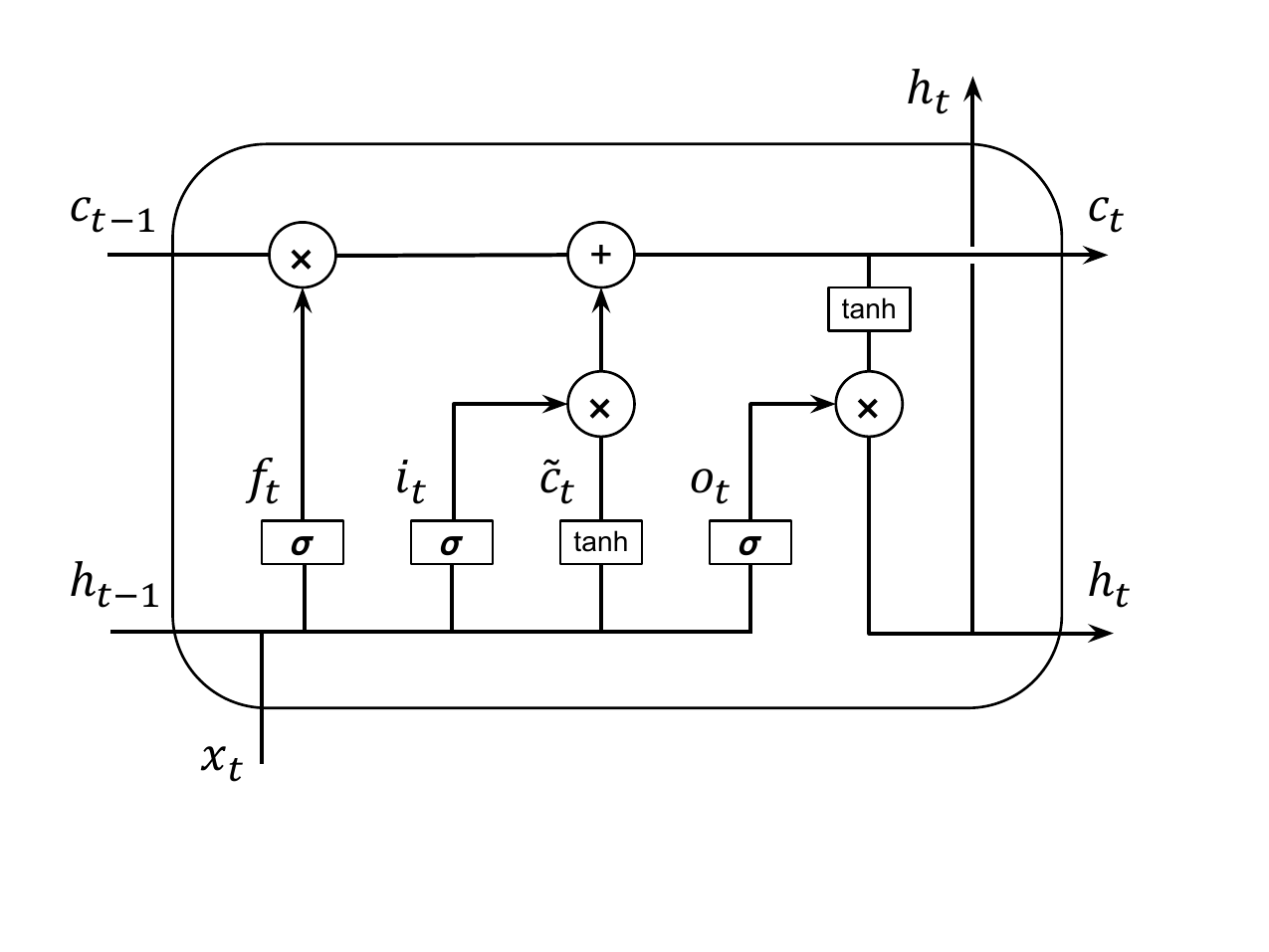}
\caption{The structure of the LSTM cell. $x$ is the input of the LSTM cell. $f$, $i$, and $o$ represent the forget, input, and output gates, respectively. $\tilde c_t$ denotes the input activation. $h$ and $c$ are hidden state and cell state, respectively. Subscript $t$ indicates the time step.}\label{fig:LSTM_cell}
\end{figure}

{\color{black}The LSTM prediction network can be easily trained with the original variables and their knockoff copies by backpropagation through time (BPTT) \cite{werbos1990backpropagation}. Similar to the training process of the LSTM autoencoder, in the case of multiple subjects (i.e., $m>1$), each subject will be treated as a batch and BPTT will be repeatedly applied to every single batch. Since a stateless LSTM layer is applied, the initial state of a new batch will not be affected by the last state of the previous batch.} 

After training this LSTM prediction network, denote by $\mathbf{z}=(z_1, z_2, \ldots, z_p)^T$ and $\tilde{\mathbf{z}}=(\tilde{z}_1, \tilde{z}_2, \ldots, \tilde{z}_p)^T$ the filter weights for the original features and their knockoff counterparts in the filter layer, respectively. Let $W_0\in \mathbb{R}^{p\times k}$ be the weight matrix for the dense layer, where $k$ is the number of neurons in the dense layer. Given the output $x_t$ in state $t$ from the dense layer, the LSTM layer (Figure \ref{fig:LSTM_cell}) first determines what old information will be discarded by the forget gate 
\begin{equation} \label{eq:LSTM_forget}
f_t=\sigma(V_fx_t+U_fh_{t-1}+b_f),
\end{equation}
where $h_{t-1}$ is the hidden state from the LSTM cell at the state $t-1$, $\sigma(\cdot)$ denotes the sigmoid function, $b_f$ is the bias parameter, and $V_f$ and $U_f$ are network weight matrices of appropriate sizes. An input gate and a cell state decide the updated values $i_t$ and propose a new candidate values $\tilde{c}_t$, respectively. These values together determine the new cell state $c_t$ as 
\begin{equation} \label{eq:LSTM_input}
\begin{aligned}
i_t&=\sigma(V_ix_t+U_ih_{t-1}+b_i), \\
\tilde{c}_t&=\text{tanh}(V_cx_t+U_ch_{t-1}+b_c),\\
c_t&=f_t\odot c_{t-1}+i_t\odot \tilde{c}_t,
\end{aligned}
\end{equation}
where $\text{tanh}(\cdot)$ denotes the hyperbolic tangent function, $\odot$ is element-wise product, $b_i$ and $b_c$ are the biases, and $V_i$, $U_i$, $V_c$, $U_c$ are network weights of appropriate sizes.
Finally, the output $o_t$ and the hidden state $h_t$ at current state $t$ can be obtained by 
\begin{equation} \label{eq:LSTM_output}
\begin{aligned}
o_t&=\sigma(V_ox_t+U_oh_{t-1}+b_o), \\
h_t&=o_t\odot \text{tanh}(c_t).
\end{aligned}
\end{equation}
Here, similarly, $b_o$ is the bias, and $V_o$ and $U_o$ are the weight matrices. 

 In Equations \eqref{eq:LSTM_forget}--\eqref{eq:LSTM_output} above, if we denote $u$ as the number of LSTM units in the LSTM layer, then weight matrices $V_{\cdot}$'s are of size $u\times k$, weight matrices $U_{\cdot}$'s are of size $u\times u$, and biases $b_{\cdot}$ are all vectors of length $u$, where  $\cdot$ in the subscript can be $f$, $i$, $c$, or $o$. 
 
 To measure the relative importance of original features and their knockoff counterparts, we track how their network weights propagate along the network through the filter layer, dense layer, LSTM layer, and output layer (for predicting $y$). Due to the complicated structure of the LSTM layer, each of the original and knockoff features make multi-path contributions to the network by going through various gates including the forget gate, input gate, cell state, and output gate. Since their contributions along these different paths (each corresponds to a gate) are likely different, we measure the relative importance of original and knockoff features when going through each of the paths separately as       

\begin{equation} \label{eq:knockoff_stats0}
\begin{aligned}
v^{(f)}&=(V_0\odot\Gamma)V_fV_1, \\
v^{(i)}&=(V_0\odot\Gamma)V_iV_1, \\
v^{(c)}&=(V_0\odot\Gamma)V_cV_1, \\
v^{(o)}&=(V_0\odot\Gamma)V_oV_1,
\end{aligned}
\end{equation}
respectively corresponding to forget gate, input gate, cell state, and output gate. Here, $V_1\in \mathbb{R}^{h}$ is the weight vector for the output layer following the LSTM layer and $\Gamma = [\gamma, \gamma, \ldots, \gamma]^T\in \mathbb{R}^{p\times k}$ with $\gamma \in \mathbb{R}^{k}$ being the weight vector for the batch normalization \cite{ioffe2015batch} between the dense layer and the LSTM layer. If the batch normalization is not used, we simply remove $\Gamma$ from the above equation. We then augment the contributions of the original feature $j$ along the four different paths into a vector $(z_jv^{(f)}_j, z_jv^{(i)}_j, z_jv^{(c)}_j, z_jv^{(o)}_j)$, and define the variable importance measure for feature  $j$ as
\begin{equation} \label{eq:knockoff_stats1}
\begin{aligned}
Z_j&=||(z_jv^{(f)}_j, z_jv^{(i)}_j, z_jv^{(c)}_j, z_jv^{(o)}_j)||_2, 
\end{aligned}
\end{equation}
where $||\cdot||_2$ denotes the $L_2$-norm.
Correspondingly, the variable importance measure for the $j$th knockoff variable is defined as 
\begin{equation*} 
\begin{aligned}
\tilde{Z}_j&=||(\tilde{z}_jv^{(f)}_j, \tilde{z}_jv^{(i)}_j, \tilde{z}_jv^{(c)}_j, \tilde{z}_jv^{(o)}_j)||_2.
\end{aligned}
\end{equation*}
Finally, the knockoff statistics are defined as
\begin{equation} \label{eq:knockoff_stats}
W_j=Z^2_j-\tilde{Z}^2_j,\ j =1,\cdots, p.
\end{equation}

With the use of our new knockoff statistics introduced in Equation \eqref{eq:knockoff_stats} above, the feature selection procedure described in Equations \eqref{eq:threshold} and \eqref{eq:threshold+} can be directly applied. It is important to note that the filter weights $z_j$ and $\tilde z_j$ need to be initialized identically when training the neural networks to ensure good feature selection performance.

Here, we have introduced how to construct knockoff statistics based on the LSTM network structure depicted in Figure \ref{fig:LSTM_prediction_nn}. Our definition of knockoff statistics can be easily adapted to other neural network architectures, including those with additional dense layers and multiple stacked LSTM layers. 

\section{Simulation studies} \label{sec:simulation}


To investigate the finite-sample capability of DeepLINK-T for FDR control in feature selection, we design various simulation models including the linear and nonlinear factor models, and different forms of the link function for the response and features for a comprehensive analysis. In all simulation designs, we set the vector of raw latent factors as  ${\mathbf f}_{t}^{raw} = (f^{raw}_{t1}, f^{raw}_{t2}, f^{raw}_{t3})^T$. Each $\mathbf{f}_{t}^{raw}$ is drawn independently from the multivariate normal distribution $N(0, \Sigma)$, where $\Sigma$ has the AR(1) structure with $(i, j)$th entry being $0.9^{|i-j|}$.  Then the factor vector $\mathbf{f}_t=(f_{t1}, f_{t2}, f_{t3})^T$ is obtained based on $\mathbf{f}^{raw}_{t}$ and Equation \eqref{eq:factor_weighted} with $w_0=0.3$ and $w_1=0.7$. Two factor models, including the linear factor model (Equation \eqref{eq:linear_fm}) and logistic factor model (Equation \eqref{eq:logistic_fm}), are used in the simulation studies to generate covariates:
\begin{equation} \label{eq:linear_fm}
\mathbf{x_t}=\boldsymbol\Lambda \mathbf{f}_t+\boldsymbol\epsilon_t,
\end{equation}
\begin{equation} \label{eq:logistic_fm}
x_{tk} = \frac{c_{k}}{1 + \exp([1, \mathbf{f}_t^T]\boldsymbol\lambda_k)} + \epsilon_{tk}, \ k = 1, \dotsb, p.
\end{equation}
Here, $c_k$, $\boldsymbol\lambda_{k}$, $\epsilon_{tk}$, and entries of $\boldsymbol\Lambda\in\mathbb{R}^{p\times 3}$ are all independently sampled from the standard normal distribution $N(0, 1)$.

The response vector $\mathbf{y}=(y_1,\ldots,y_n)^T$ is simulated from Equation \eqref{eq:regression_model}, and both the linear link function (Equation \eqref{eq:linear_h}) and nonlinear link function (Equation \eqref{eq:nonlinear_h}) are considered in our simulation study:

\begin{equation} \label{eq:linear_h}
l(\mathbf{x_t})=\mathbf{x_t}\boldsymbol\beta,
\end{equation}
\begin{equation} \label{eq:nonlinear_h}
l(\mathbf{x_t})=\sin(\mathbf{x_t}\boldsymbol\beta)\exp(\mathbf{x_t}\boldsymbol\beta),
\end{equation}
where $\boldsymbol\beta=(\beta_1,\ldots,\beta_p)^T$ is the coefficient vector with each $\beta_k, 1\leq k\leq p$, being either $A$, $-A$, or $0$. Here, $A$ is some positive value standing for the signal amplitude. We randomly pick $s$ components from $\boldsymbol\beta$ and set them to be $A$ or $-A$ with the same probability; the features corresponding to these $s$ components are important features and expected to be selected by a good feature selection method. The remaining $p-s$ components of $\boldsymbol\beta$ are then set to be $0$ indicating that the corresponding features are unimportant. With the linear link function, it is evident that a large value of $A$ indicates a strong signal strength of the corresponding feature. However, for nonlinear link functions, it is possible that $A$ no longer has a monotonic increasing relationship with the signal strength. Indeed, there does not exist a commonly accepted measure for the signal strength in nonlinear model settings; see \cite{2021DeepLINK} for more detailed discussions on this.

Furthermore, to verify the capability of DeepLINK-T to tackle potential model misspecification in real-world applications, we test DeepLINK-T on the response relationship with latent confounders using the following model
\begin{equation} \label{eq:latent_confounder}
 y_{t} = l(\mathbf{x}_{t}) + \frac{1}{3}(f_{t1}+f_{t2}+f_{t3}) + \varepsilon_{it},\ t = 1, \cdots, n.
\end{equation}
Note that although $\mathbf f_t$ can be viewed as a ``function" of both $\mathbf x_t$ and the idiosyncratic error $\mathbf \epsilon_t$,  the dependence of $y_t$ on $\mathbf x_t$ in \eqref{eq:latent_confounder} above does not admit the functional form defined in \eqref{eq:regression_model}. In this sense, \eqref{eq:regression_model} is a misspecified model when data is generated from \eqref{eq:latent_confounder}.  

Sections \ref{new.sec3.2} and \ref{new.sec3.3} consider the case of $m=1$, while Section \ref{new.sec3.4} investigates the case of $m>1$.

\begin{figure}[t]%
\centering
\includegraphics[width=\textwidth]{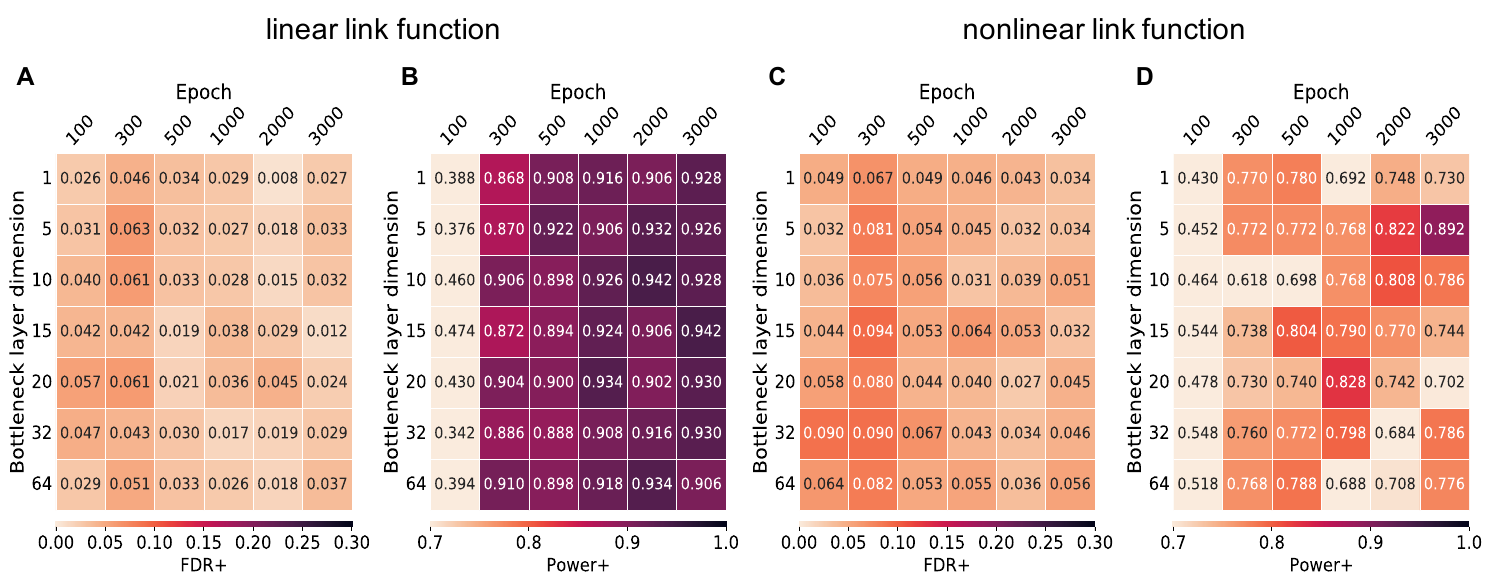}
\caption{The impacts of bottleneck dimensionality and training epochs on DeepLINK-T using the \textbf{linear} factor model. A and B are FDR and power under the setting of linear link function. C and D are FDR and power under the setting of nonlinear link function. The number of training epochs of both LSTM autoencoder and LSTM prediction network is specified on the x-axis. The bottleneck dimensionality of the LSTM autoencoder is specified on the y-axis. The pre-specified FDR level is $q=0.2$.}\label{fig:heatmap_fm_linear}
\end{figure}

\begin{figure}[t]%
\centering
\includegraphics[width=\textwidth]{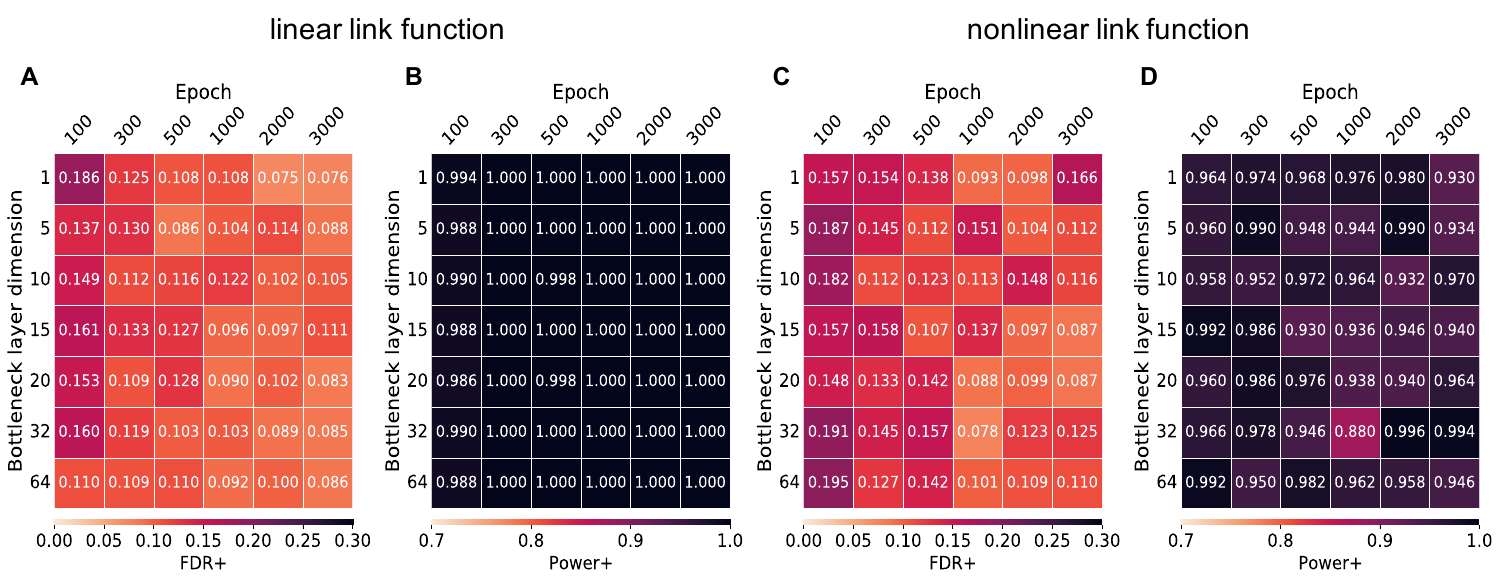}
\caption{The impacts of bottleneck dimensionality and training epochs on DeepLINK-T using the \textbf{logistic} factor model. A and B are FDR and power under the setting of linear link function. C and D are FDR and power under the setting of nonlinear link function. The number of training epochs of both LSTM autoencoder and LSTM prediction network is specified on the x-axis. The bottleneck dimensionality of the LSTM autoencoder is specified on the y-axis. The pre-specified FDR level is $q=0.2$.}\label{fig:heatmap_fm_nonlinear}
\end{figure}

\begin{figure}[t]%
\centering
\includegraphics[width=\textwidth]{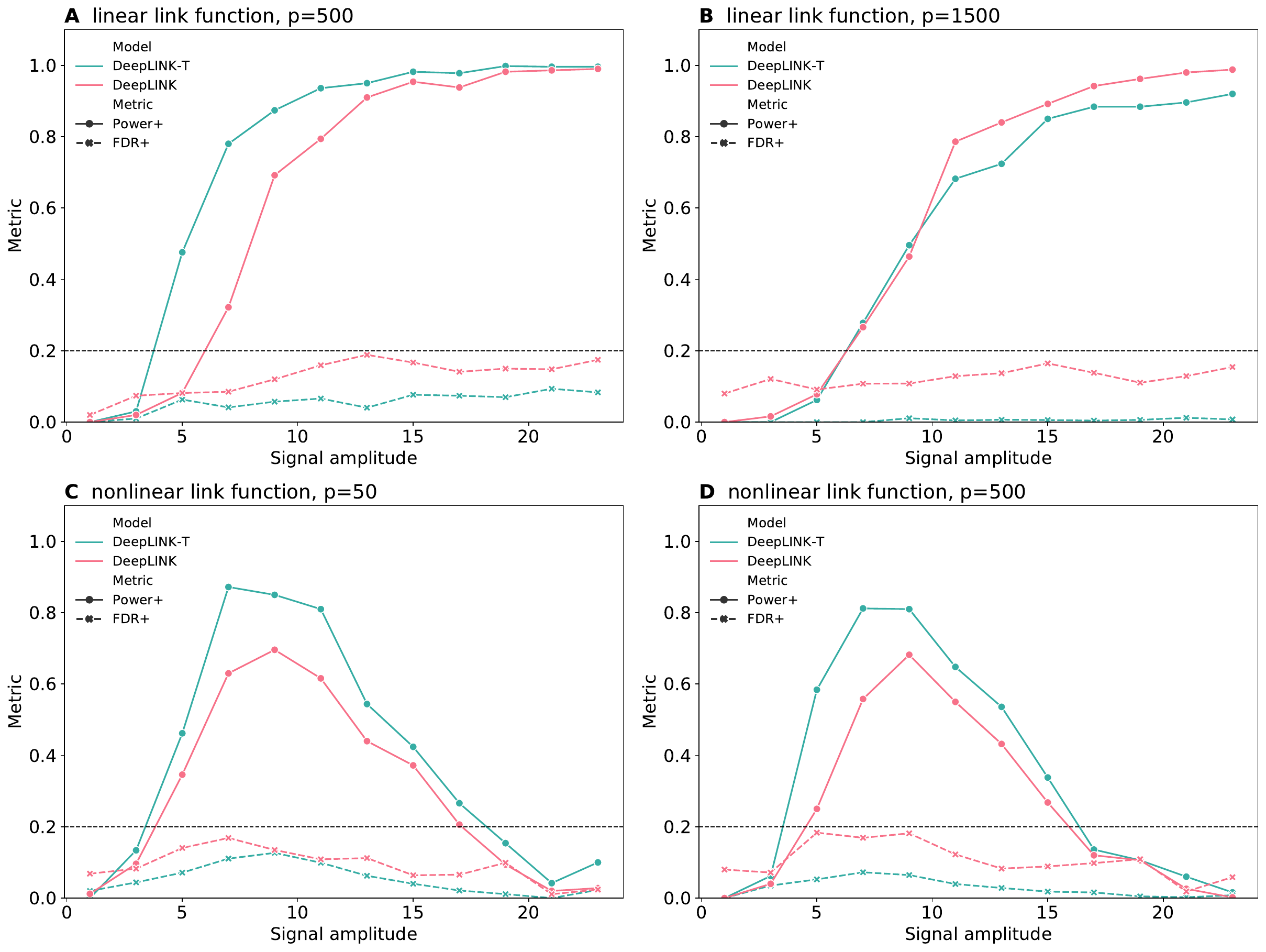}
\caption{Comparisons of DeepLINK-T and DeepLINK based on simulated data with the \textbf{linear factor model}. The number of subjects $m$ is 1, the number of time points $n$ is 1000, and the number of true signals is 10. Each subtitle suggests the link function $l$ and the number of features $p$. The solid lines and dashed lines stand for Power+ and FDR+, respectively. The green lines and red lines represent the results of DeepLINK-T and DeepLINK, respectively.}\label{fig:lineplot_linear_h}
\end{figure}

\begin{figure}[t]%
\centering
\includegraphics[width=\textwidth]{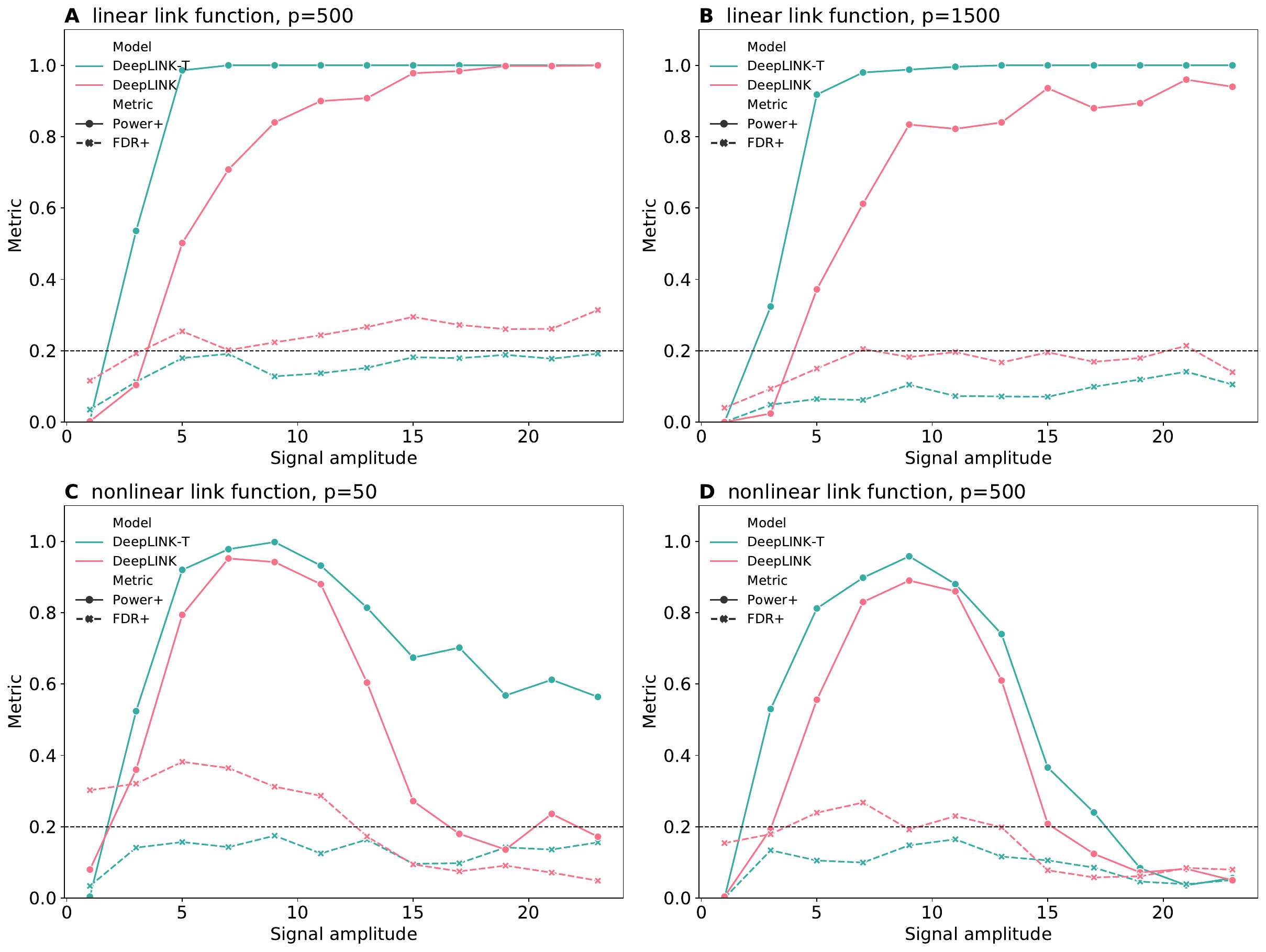}
\caption{Comparisons of DeepLINK-T and DeepLINK based on simulated data with the \textbf{logistic factor model}. The number of subjects $m$ is 1, the number of time points is 1000, and the number of true signals is 10. Each subtitle suggests the link function $l$ and the number of features $p$. The solid lines and dashed lines stand for Power+ and FDR+, respectively. The green lines and red lines represent the results of DeepLINK-T and DeepLINK, respectively.}\label{fig:lineplot_logistic_h}
\end{figure}

\begin{figure}[t]%
\centering
\includegraphics[width=\textwidth]{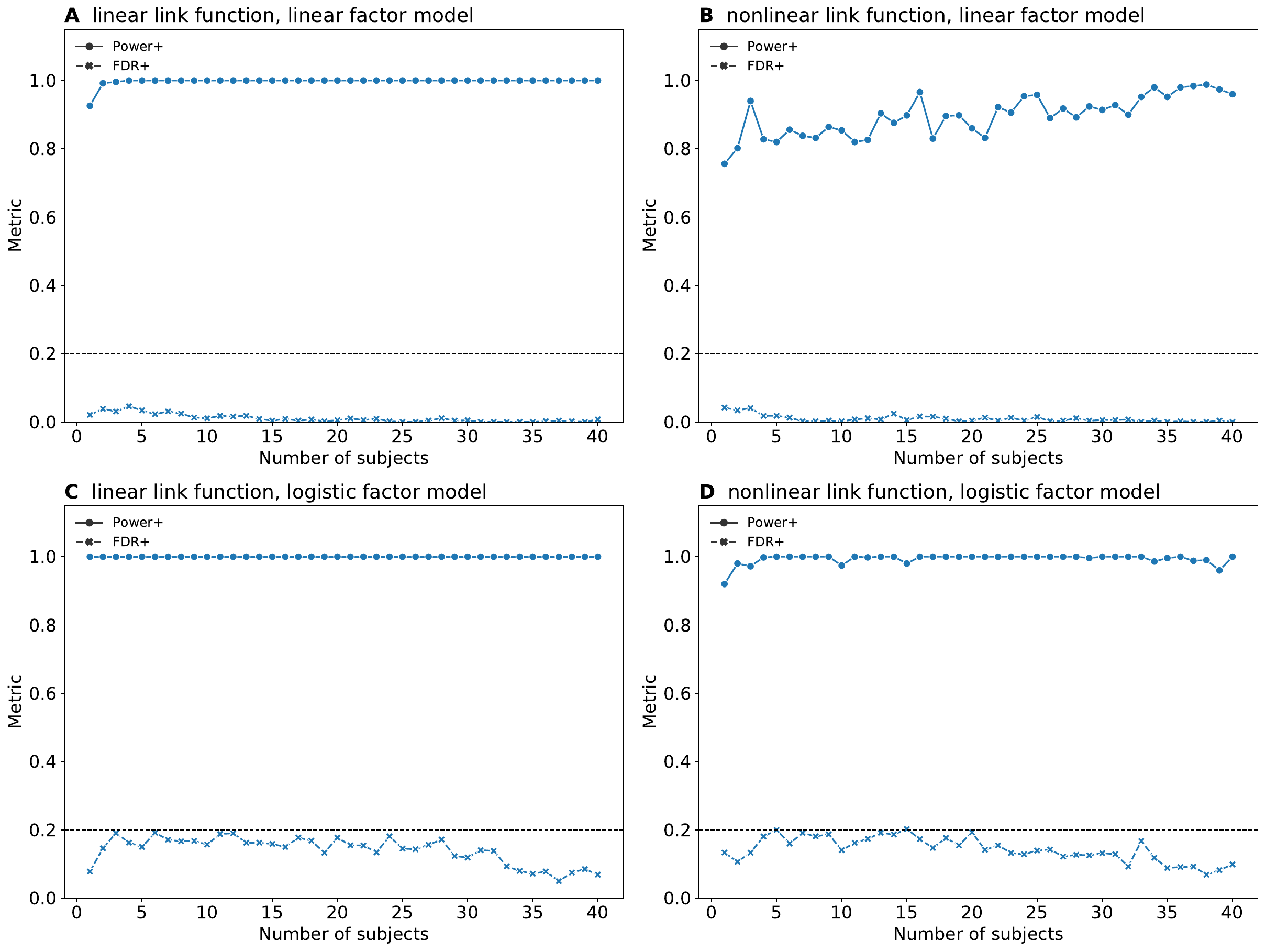}
\caption{The impacts of the number of subjects $m$ on DeepLINK-T. In this simulation, the number of time points is 1000, the number of features is 500, and the number of true signals is 10. Each subtitle suggests the link function $l$ and the factor model. The solid lines and dashed lines stand for Power+ and FDR+, respectively.}\label{fig:lineplot_subjects}
\end{figure}

\subsection{The impacts of hyperparameters and model misspecification on DeepLINK-T} \label{new.sec3.2}

We fixed the number of subjects $m=1$, the number of time points $n=1000$, the number of features $p=500$, the signal amplitude $A=10$, and the pre-specified FDR level $q=0.2$ to investigate the effects of hyperparameters including bottleneck dimensionality in the LSTM autoencoder and the training epochs for both networks (i.e., LSTM autoencoder and LSTM prediction network) in DeepLINK-T on the knockoffs inference results.  Figures \ref{fig:heatmap_fm_linear} and \ref{fig:heatmap_fm_nonlinear} demonstrate the performance of DeepLINK-T in the linear factor model and logistic factor model, respectively. Sub-figures A and B present the FDR+ and Power+ for the linear link function setting when varying the values for bottleneck dimensionality and training epochs. Sub-figures C and D present the FDR+ and Power+ in the setting of nonlinear link function. These results reveal lower power when the number of training epochs is less than 300, indicating the potential underfitting status of the networks. Another observation is that DeepLINK-T is robust to the choice of bottleneck dimensionality. The FDR stays under control and the power is stable as the bottleneck dimensionality varies from 1 to 64 when the number of training epochs exceeds 500. The underestimation (bottleneck dimensionality $<$ 3) and overestimation (bottleneck dimensionality $>$ 3) of the number of latent factors do not substantially impact the FDR and power, justifying the robustness of DeepLINK-T under potential model misspecification. 

Next, we introduced latent confounders as described in Equation \eqref{eq:latent_confounder} into the data generating models, aiming to  assess the ability of DeepLINK-T in addressing model misspecification more comprehensively. Supplementary Figures S1 and S2 present the FDR+ and power+ metrics based on the linear factor model and logistic factor model, respectively. The results reveal that DeepLINK-T, when trained with fewer than 500 epochs, exhibits reduced power with inflated FDR, particularly in the case of the logistic factor model. With larger training epochs exceeding 1000, the performance of DeepLINK-T becomes comparable to simulations without latent confounders. Additionally, the bottleneck dimensionality does not meaningfully influence DeepLINK-T's performance. All together, these results suggest that DeepLINK-T demonstrates robustness under the scenario of model misspecification.

\subsection{Comparisons of DeepLINK-T and DeepLINK} \label{new.sec3.3}

To further investigate the ability of DeepLINK-T to handle longitudinal data, we applied both DeepLINK-T and DeepLINK \cite{2021DeepLINK} to various simulated data sets for a comprehensive comparison. Since DeepLINK-T is robust to different values of bottleneck dimensionality and 1000 training epochs are enough for the learned model to converge, we fixed the bottleneck dimensionality to be 15. The number of training epochs was set to 1000 for both DeepLINK-T and DeepLINK in this subsection. The pre-specified FDR level was again set to 0.2. We tested both algorithms on the data simulated with the combination of different link functions and different factor models listed in the simulation designs. For each simulation setting, the two algorithms were tested on 50 independently simulated data sets. Then the average FDP and TDP were calculated as the empirical FDR and power, respectively. The comparison results are shown in Figures \ref{fig:lineplot_linear_h} and \ref{fig:lineplot_logistic_h}. 

Figure \ref{fig:lineplot_linear_h} shows the comparison based on the linear factor model. Both DeepLINK and DeepLINK-T effectively control the FDR under the target level $0.2$. Yet, DeepLINK-T successfully maintains the FDR at a lower level while achieving a higher power compared to DeepLINK in all but the setting in panel B, where the power of DeepLINK-T is only slightly lower than DeepLINK when the signal amplitude is high. 

The simulation results based on the logistic factor model (Figure \ref{fig:lineplot_logistic_h}) provide additional evidence of DeepLINK-T's superior capability of FDR control in feature selection.  DeepLINK can no longer control the FDR in the setting of the nonlinear factor model, while DeepLINK-T can still control the FDR under the target level and simultaneously outperform DeepLINK in terms of power. An interesting observation is that the power curve is inverted-U shaped as the signal amplitude $A$ increases when the nonlinear link function $l$ is applied, which suggests that the signal amplitude may not be an appropriate measure for the signal strength in certain nonlinear models. We remark that similar phenomenon on the power curve was observed in \cite{2021DeepLINK}. Overall, these simulation results highlight the power of DeepLINK-T in dealing with longitudinal data. The simulation settings involving different factor models and link functions demonstrate the ability of DeepLINK-T in coping with complex models, indicating its great potential in real applications. 

\textcolor{black}{
\subsection{The impacts of number of subjects on DeepLINK-T} \label{new.sec3.4}
Another advanced feature of DeepLINK-T lies in its ability to handle multiple subjects (time series), which is prevalent in real-world applications. An illustrative example is the gut microbiome metagenomic time series, where each time series may independently correspond to a different individual. Hence, in this simulation case, we investigate the influence of the number of subjects on the performance of DeepLINK-T. We maintained a fixed number of time points $n=1000$, number of features $p=500$, signal amplitude $A=10$, and pre-specified FDR level $q=0.2$. Multiple subjects were independently and identically simulated using the procedure described above. Figure \ref{fig:lineplot_subjects} displays the performance variation as the number of subjects changes across different factor models and link functions. The results consistently demonstrate that FDR+ decreases as the number of subjects increases. Moreover, in simulations involving the nonlinear link function and the linear factor model, power+ increases with the growing number of subjects. In summary, these findings underscore that the generalization of variance estimation outlined in Equation \eqref{eq:normal_variance_est} from a single subject to multiple subjects is reasonable. This further substantiates the prospect of DeepLINK-T in handling complicated real-world data sets.
}

\section{Real data applications} \label{new.sec4}

We now apply DeepLINK-T to three real-world metagenomic time series data sets to further validate its practical performance. All metagenomic data was processed by a centered log ratio (CLR) transformation. For all three applications, the numbers of training epochs and the bottleneck dimensionality were set to 1000 and 15, respectively. The pre-specified FDR level was chosen as 0.2. In order to alleviate the randomness during the weight initialization and the training process of the neural networks, we repeatedly implemented DeepLINK-T on each data set 200 times.

\subsection{Application to longitudinal gut microbiome data of early infants} \label{new.sec4.1}

First, we applied DeepLINK-T to microbial longitudinal samples from early infants \cite{bokulich2016antibiotics, velten2022identifying}. Our focus was on identifying important bacterial genera associated with the abundance levels of \textit{Bacteroides} and \textit{Bifidobacterium}, which are predominant genera in the stool of early life from the facultative aerobic \textit{Enterobacteriaceae} family. These two genera were found to be gradually displaced by a diverse mixture of \textit{Clostridiales} in two years \cite{bokulich2016antibiotics}. Since we used the abundance of one genus among all the genera as the response, a direct application of the CLR transformation to all genera could cause collinearity between the explanatory and response variables. Instead, a modified CLR transformation was adopted to address the issue. For the explanatory genera, we directly applied CLR 
\begin{equation} \label{eq:modified_clr1}
\textrm{CLR}(x_i)=\ln x_i-\frac{1}{D}\sum_{j=1}^{D}\ln x_j,
\end{equation}
where $D$ is the number of explanatory genera and $x_i$ denotes the composition part of genus $i$. Then we transformed the count of response genus $y$ using

\begin{equation} \label{eq:modified_clr2}
\hat{y} = \ln y-\frac{1}{D}\sum_{j=1}^{D}\ln x_j.
\end{equation}

Considering that the data is rather sparse, we filtered out samples with over 50\% missing time points. At the feature domain, genera that are absent in over 90\% samples were also filtered out. Consequently, the processed data set is comprised of 36 subjects, each sequenced monthly over a 24-month period. Fifty common genera were identified among all samples. Hence, the data array is of size $m \times n \times p = 36\times24\times50$.

\begin{table}[t]
\tiny
\begin{tabularx}{0.98\textwidth}{lclc}
\toprule
         Genus selected for \textit{Bacteroides} &  Frequency & Genus selected for \textit{Bifidobacterium} &  Frequency \\
\midrule
   f\_\_Porphyromonadaceae; g\_\_Parabacteroides &        200  &    f\_\_Clostridiaceae; g\_\_SMB53 &        196  \\
         f\_\_Lachnospiraceae; g\_\_Epulopiscium &        198 &    f\_\_; g\_\_ &        186       \\
          f\_\_Veillonellaceae; g\_\_Megasphaera &        192 &     f\_\_Ruminococcaceae; g\_\_ &        185      \\
f\_\_Veillonellaceae; g\_\_Phascolarctobacterium &        182 &     f\_\_Micrococcaceae; g\_\_Rothia &        185      \\
       f\_\_Fusobacteriaceae; g\_\_Fusobacterium &        182 &   f\_\_Lactobacillaceae; g\_\_Lactobacillus &        174 \\
                f\_\_Micrococcaceae; g\_\_Rothia &        170 &   f\_\_Ruminococcaceae; g\_\_Anaerotruncus &        163   \\
                 f\_\_Erysipelotrichaceae; g\_\_ &        168 &     f\_\_Erysipelotrichaceae; g\_\_[Eubacterium] &        155        \\
            f\_\_Prevotellaceae; g\_\_Prevotella &        162 &   f\_\_Lachnospiraceae; g\_\_[Ruminococcus] &        153                  \\
          f\_\_Lachnospiraceae; g\_\_Coprococcus &        161 &   f\_\_Erysipelotrichaceae; g\_\_ &        152          \\
                 f\_\_Clostridiaceae; g\_\_SMB53 &        151 &   f\_\_Clostridiaceae; g\_\_Clostridium &        147    \\
   f\_\_Bifidobacteriaceae; g\_\_Bifidobacterium &        146 & f\_\_Turicibacteraceae; g\_\_Turicibacter &        146 \\
         f\_\_Ruminococcaceae; g\_\_Ruminococcus &        137 & f\_\_Clostridiaceae; g\_\_ &        144 \\
         f\_\_Actinomycetaceae; g\_\_Actinomyces &        129 & f\_\_Coriobacteriaceae; g\_\_Eggerthella &        144 \\
         f\_\_Ruminococcaceae; g\_\_Oscillospira &        124 & f\_\_Peptostreptococcaceae; g\_\_[Clostridium] &        131 \\
  f\_\_Peptostreptococcaceae; g\_\_[Clostridium] &        120 & f\_\_Veillonellaceae; g\_\_Veillonella &        130 \\
    f\_\_Erysipelotrichaceae; g\_\_Coprobacillus &        120 & f\_\_Erysipelotrichaceae; g\_\_Coprobacillus &        120 \\
          f\_\_Pasteurellaceae; g\_\_Haemophilus &        106 & f\_\_Moraxellaceae; g\_\_Acinetobacter &        115 \\
                   f\_\_Coriobacteriaceae; g\_\_ &        105 & f\_\_Staphylococcaceae; g\_\_Staphylococcus &        109 \\
           f\_\_Clostridiaceae; g\_\_Clostridium &         98 & f\_\_Fusobacteriaceae; g\_\_Fusobacterium &        108 \\
     f\_\_Staphylococcaceae; g\_\_Staphylococcus &         98 & f\_\_Bacteroidaceae; g\_\_Bacteroides &        102 \\ 
\bottomrule
\end{tabularx}
\caption{Selection frequencies of top 20 selected genera for \textit{Bacteroides} and \textit{Bifidobacterium} over 200 runs.}
\label{ex1_sclr_table}
\end{table}

Supplementary Figure S3 illustrates the histogram of selection frequencies by DeepLINK-T over 200 runs. The top 20 microbial genera, along with their selected frequencies, are detailed in Table \ref{ex1_sclr_table}. The abundance variations of \textit{Bacteroides} and \textit{Bifidobacterium} with their top 10 selection genera are shown in Supplementary Figures S4 and S5, respectively. These findings showcase the consistent ability of DeepLINK-T to reliably identify the most relevant genera associated with \textit{Bacteroides} and \textit{Bifidobacterium} across the 200 runs. 

Specifically, \textit{Parabacteroides}, identified as the top significant genus associated with \textit{Bacteroides} in the gut of early infants (Table \ref{ex1_sclr_table} and Supplementary Figure S4), constitutes one of the major groups of bacteria that inhabit the human gastrointestinal tract, substantially influencing the microbial diversity and composition \cite{bokulich2016antibiotics} there. For \textit{Bifidobacterium}, \textit{Rothia} (Table \ref{ex1_sclr_table} and Supplementary Figure S5) was found to be positively correlated with \textit{Bifidobacterium} \cite{fehr2020breastmilk}. Both \textit{Rothia} and \textit{Bifidobacterium} are integral components of breast milk, playing crucial roles in maintaining the immune homeostasis of infant gut \cite{fehr2020breastmilk, arrieta2015early}.

\subsection{Application to marine metagenomic time series data} \label{new.sec4.2}

\subsubsection{Identifying primary chlorophyll-a producer}\label{new.sec4.2.1}

Next, we applied DeepLINK-T to analyze the San Pedro Ocean time series (SPOT) data set \cite{cram2015seasonal, yeh2022contrasting}. The ocean's surface harbors microscopic organisms known as phytoplankton, which are primary producers utilizing pigments like chlorophyll-a (chl-a) for photosynthesis. The quantification of chlorophyll in the surface water serves as a reliable indicator of primary production levels in the ocean's upper layer \cite{sigman2012biological}. Consequently, our objective is to explore the effect of the abundance of chloroplast on the chlorophyll-a concentration.

The SPOT data set consists of both 16S rRNA data and 18S rRNA data collected from San Pedro Channel, situated off the coast of Los Angeles, spanning the period from August 2000 to July 2018. Samples were taken from different depths including 5m, 150m, 500m, and etc. Specifically, we used samples from 5m to ensure the maximum number of time points and capture the most abundant microbiome. After aligning the common time point between 16S and the metadata, we identified 161 time points in total. The analysis was conducted at the order, family, and genus levels. A taxon was filtered out if it is missing in over 90\% of time points to ensure the robustness. Finally, the data arrays (matrices) are of sizes $1\times161\times88$, $1\times161\times70$, and $1\times161\times56$ at the genus, family, and order levels, respectively. The CLR normalization was applied to compute the relative abundance of the explanatory variables.

DeepLINK-T again demonstrated its efficacy in discovering taxa closely associated with the concentration of chl-a. The histogram of selection frequencies by DeepLINK-T on the SPOT data set is depicted in Supplementary Figure S6. Supplementary Figure S7 further delineates the relationship between chl-a concentration and the abundance of its top three selected taxa. Notably, peak chl-a concentration aligns with the peak abundance of \textit{Heterosigma}, \textit{Pyramimonas}, and \textit{Corethrales}. Conversely, the peak chl-a concentration coincides with the troughs in the abundance of \textit{Chlorarachniaceae} family with its parent \textit{Chlorarachniophyceae} order.

\begin{table}[t]
\begin{tabular}{lc}
\toprule
                                          Genus &  Frequency \\
\midrule
              f\_\_Chattonellaceae;g\_\_Heterosigma &         51 \\
             f\_\_Pyramimonadaceae;g\_\_Pyramimonas &         51 \\
            f\_\_Rhizosoleniaceae;g\_\_Rhizosolenia &         46 \\
          f\_\_Chlorarachniaceae;g\_\_Partenskyella &         34 \\
                   f\_\_Corethraceae;g\_\_Corethron &         25 \\
   f\_\_Prasino-clade-9\_XX;g\_\_Prasino-clade-9\_XXX &         19 \\
f\_\_Sarcinochrysidaceae;g\_\_Sarcinochrysidaceae\_X &         19 \\
 f\_\_Prasino-clade-7\_X-B;g\_\_Prasino-clade-7\_X-B2 &         18 \\
          f\_\_Thalassiosiraceae;g\_\_Thalassiosira &         16 \\
    f\_\_Chrysochromulinaceae;g\_\_Chrysochromulina &         14 \\
\midrule
                                                 Family &  Frequency \\
\midrule
       o\_\_Chlorarachniophyceae;f\_\_Chlorarachniaceae &         44 \\
                     o\_\_Corethrales;f\_\_Corethraceae &         41 \\
               o\_\_Chattonellales;f\_\_Chattonellaceae &         38 \\
           o\_\_Prasino-clade-9;f\_\_Prasino-clade-9\_XX &         22 \\
o\_\_Coscinodiscophyceae\_X;f\_\_Coscinodiscophyceae\_XXX &         21 \\
          o\_\_Prasino-clade-7;f\_\_Prasino-clade-7\_X-B &         17 \\
             o\_\_Bolidomonadales;f\_\_Bolidomonadaceae &         14 \\
             o\_\_Rhizosoleniales;f\_\_Rhizosoleniaceae &         14 \\
    o\_\_Prymnesiophyceae\_XX;f\_\_Prymnesiophyceae\_XXXX &         11 \\
                   o\_\_Eutreptiales;f\_\_Eutreptiaceae &         10 \\

\midrule
                                          Order &  Frequency \\
\midrule
c\_\_Filosa-Chlorarachnea;o\_\_Chlorarachniophyceae &         48 \\
           c\_\_Prasinophyceae;o\_\_Prasino-clade-9 &         34 \\
              c\_\_Bacillariophyta;o\_\_Corethrales &         29 \\
            c\_\_Raphidophyceae;o\_\_Chattonellales &         29 \\
    c\_\_Bacillariophyta;o\_\_Coscinodiscophyceae\_X &         27 \\
           c\_\_Prasinophyceae;o\_\_Prasino-clade-7 &         26 \\
            c\_\_Bolidophyceae;o\_\_Bolidomonadales &         20 \\
          c\_\_Bacillariophyta;o\_\_Rhizosoleniales &         15 \\
                c\_\_Pavlovophyceae;o\_\_Pavlovales &         10 \\
           c\_\_Prasinophyceae;o\_\_Pyramimonadales &          9 \\
\bottomrule
\end{tabular}
\caption{Selection frequencies of top 10 selected taxa for chl-a concentration at the genus, family, and order levels over 200 runs.}
\label{ex2_chla_table}
\end{table}

Table \ref{ex2_chla_table} presents the top selected taxa along with their corresponding selection frequencies. Among the top selected genera, Jackson et al. \cite{jackson1990contribution} observed that the deep chlorophyll maximum (DCM) is primarily influenced by a species from the \textit{Rhizosolenia} genus. Additionally, Zhao et al. \cite{zhao2010relation} identified a significant correlation between the concentration of chl-a and the top two selected genera \textit{Pyramimonas} and \textit{Heterosigma}.

\textcolor{black}{
\subsubsection{Identifying taxa significantly associated with prokaryotic production}\label{new.sec4.2.2}
Similar to chlorophyll-a concentration, the leucine incorporation rate serves as a crucial metric for measuring prokaryotic production \cite{kirchman1985leucine}. Employing a similar approach, we investigated the relationship between prokaryotic abundance levels and the leucine incorporation rate. Following data filtration and CLR normalization, the prokaryotic abundance data matrices are of sizes $1\times161\times270$, $1\times161\times173$, and $1\times161\times110$ at the genus, family, and order levels, respectively.}

\textcolor{black}{
Supplementary Figures S8 and S9 display the selection frequencies and the relationship between leucine incorporation rate and the abundance levels of its top three selected prokaryotes. Notably, the peak leucine incorporation rate is in line with the peak abundance of \textit{Tenacibaculum}, \textit{Planktomarina}, \textit{Flavobacteriaceae}, \textit{Halieaceae}, and \textit{Cellvibrionales}. Conversely, the peak leucine incorporation rate coincides with the troughs in the abundance of \textit{Defluviicoccales} order. Table S1 presents the top selected taxa with their selection frequencies. Among the selection taxa, \textit{Rhodobacteraceae} family and orders from \textit{Alphaproteobacteria} class, previously identified by Lamy et al. \cite{lamy2009temporal} and Wemheuer et al. \cite{wemheuer2015green}, were confirmed to significantly  contribute to the leucine incorporation rate during the peak of an iron-fertilized phytoplankton bloom in the ocean. Moreover, the \textit{Polaribacter} genus was also found as a predominant contributor to both leucine incorporation and prokaryotic abundance in regions characterized by high-nutrient, low-chlorophyll conditions in the upper 100 meters of the ocean \cite{obernosterer2011distinct}.
}

\begin{table}[t]
\begin{tabular}{lc}
\toprule
                                  Genus &  Frequency \\
\midrule
   f\_\_Oscillospiraceae;g\_\_Oscillibacter &        172 \\
            f\_\_Micrococcaceae;g\_\_Rothia &        159 \\
 f\_\_Campylobacteraceae;g\_\_Campylobacter &        133 \\
 f\_\_Staphylococcaceae;g\_\_Staphylococcus &         99 \\
        f\_\_Spirochaetaceae;g\_\_Treponema &         54 \\
     f\_\_Odoribacteraceae;g\_\_Odoribacter &         31 \\
    f\_\_Ruminococcaceae;g\_\_Angelakisella &         30 \\
                  f\_\_;g\_\_Intestinimonas &         28 \\
 f\_\_Carnobacteriaceae;g\_\_Granulicatella &         21 \\
f\_\_Ruminococcaceae;g\_\_Ruminiclostridium &         18 \\
\midrule
                                    Family &  Frequency \\
\midrule
      o\_\_Clostridiales;f\_\_Oscillospiraceae &        160 \\
o\_\_Campylobacterales;f\_\_Campylobacteraceae &        124 \\
        o\_\_Bacillales;f\_\_Staphylococcaceae &         91 \\
      o\_\_Spirochaetales;f\_\_Spirochaetaceae &         79 \\
        o\_\_Micrococcales;f\_\_Micrococcaceae &         76 \\
      o\_\_Bacteroidales;f\_\_Odoribacteraceae &         63 \\
        o\_\_Synergistales;f\_\_Synergistaceae &         58 \\
      o\_\_Brachyspirales;f\_\_Brachyspiraceae &         58 \\
    o\_\_Mycoplasmatales;f\_\_Mycoplasmataceae &         42 \\
 o\_\_Enterobacterales;f\_\_Enterobacteriaceae &         28 \\
\midrule
                                        Order &  Frequency \\
\midrule
c\_\_Epsilonproteobacteria;o\_\_Campylobacterales &        154 \\
            c\_\_Spirochaetia;o\_\_Brachyspirales &        116 \\
              c\_\_Synergistia;o\_\_Synergistales &         98 \\
                c\_\_Chlamydiia;o\_\_Chlamydiales &         93 \\
             c\_\_Mollicutes;o\_\_Mycoplasmatales &         89 \\
      c\_\_Gammaproteobacteria;o\_\_Aeromonadales &         59 \\
  c\_\_Gammaproteobacteria;o\_\_Oceanospirillales &         57 \\
        c\_\_Actinobacteria;o\_\_Streptomycetales &         39 \\
        c\_\_Gammaproteobacteria;o\_\_Vibrionales &         33 \\
     c\_\_Actinobacteria;o\_\_Propionibacteriales &         31 \\
\bottomrule
\end{tabular}
\caption{Selection frequencies of top 10 selected taxa for differentiating the FOS and PDX groups at the genus, family, and order levels over 200 runs.}
\label{ex3_treatment_table}
\end{table}

\subsection{Application to dietary glycans treatment time series data} \label{new.sec4.3}

In our final exploration, we applied DeepLINK-T to a time series data set involving dietary glycans treatment \cite{creswell2020high}, which encompasses shotgun data collected from the stool samples of 61 healthy volunteers at dense temporal resolution (about 4 times per week over 10 weeks). The data set was used to study two different glycans (fructooligosaccharides (FOS) and polydextrose (PDX)). To discern the important bacterial groups associated with the two treatments, the data set was aggregated at various taxonomic levels, including genus, family, and order. The CLR was used to transform the count data into relative abundance. After filtering out samples with over 50\% missing data and taxa absent in over 90\% subjects, 25 subjects for the FOS group and 27 subjects for the PDX group were retained. Each sample accounted for 40 time points, revealing a data set comprising 241 genera, 77 families, and 38 orders, respectively. Therefore, the input data arrays are of sizes $52\times40\times241$, $52\times40\times77$, and $52\times40\times38$ at the three levels, respectively.

The distribution of selection frequencies, depicted in Supplementary Figure S10, underscores the robustness of DeepLINK-T to stably identify taxa importantly associated with the two treatments across different taxonomic levels. The pertinent taxa identified by DeepLINK-T are shown in Table \ref{ex3_treatment_table}. Consistent with the finding from Creswell et al. \cite{creswell2020high}, \textit{Ruminiclostridum} and \textit{Odoribacter} are significantly correlated with the PDX treatment, often resulting in an increase in a more diverse set of bacterial genera. Furthermore, FOS treatment is associated with a reduction in the proportion of the \textit{Rothia} genus \cite{azagra2019oligosaccharides}. Supplementary Figures S11--S13 additionally elucidate that DeepLINK-T effectively identifies taxa such as \textit{Odoribacter}, \textit{Angelakisella}, and \textit{Ruminiclostridium} exhibiting decreased abundant in both groups. Concurrently, \textit{Campylobacteraceae} displays an increase in abundance in the FOS group and a decrease in the PDX group throughout the treatment period.

\section{Discussions}\label{sec12}
In this paper, we have introduced a new method, DeepLINK-T, designed for stable feature selection in deep learning applications with longitudinal time series data. DeepLINK-T leverages the LSTM autoencoder based on the latent factor assumption to generate time series knockoff variables under the potentially strong dependency structure. Then another LSTM prediction network is trained on the combination of the original variables and knockoff variables to construct the knockoff statistics for knockoffs inference. Our method extends the application of the original model-X knockoffs framework to the recurrent neural networks (RNNs) and temporal data. Through comprehensive simulation studies, we have demonstrated that DeepLINK-T attains high power and successfully controls the FDR at a desired level in terms of feature selection. Notably, DeepLINK-T outperforms its predecessor, DeepLINK, by exploiting the serial dependency information. Moreover, our empirical analyses highlight the utility and effectiveness of DeepLINK-T in practice.

Future advancements in this methodology could explore the integration of the Transformer \cite{vaswani2017attention} instead of the LSTM in both the autoencoder and the prediction networks for generating the knockoff variables and constructing the knockoff statistics. The attention mechanism in the Transformer could be helpful to learn long-term serial dependency within the longitudinal data. Additionally, Transformer is faster than LSTM in both training and inference phases due to its enhanced parallelizability. Furthermore, the model-X knockoffs framework can be conceptualized as a general inference machine, where knockoff statistics play a crucial role. In our current work, we have utilized the method proposed by \cite{2018MXK} to calculate these statistics. Future endeavors could explore innovative approaches to constructing new knockoff statistics that can yield higher power with effective FDR control. Finally, it is also interesting to integrate the two underlying deep learning networks into one unified deep learning model for end-to-end training and inference with flexibility, interpretability, and stability. This consolidation would enhance the tool's user-friendliness and overall efficiency.

\backmatter

\bmhead{Supplementary information}
The Supplementary Material contains additional simulation and real data results from Sections \ref{sec:simulation} and \ref{new.sec4}, respectively.

\bmhead{Acknowledgments}
This work was supported by NSF Grants EF-2125142 and DMS-2324490.

\section*{Declarations}

\subsection*{Competing interests}
The authors declare no competing interests.

\subsection*{Availability of data and materials}
The study contains three publicly available time series metagenomic data sets, including a longitudinal gut microbiome data set \cite{bokulich2016antibiotics, velten2022identifying}, a marine metagenomic time series data set \cite{yeh2022contrasting}, and a dietary glycans treatment time series data set \cite{creswell2020high}. The data can also be found in the Github repository for DeepLINK-T (\url{https://github.com/zuowx/DeepLINK-T}).

\subsection*{Code availability}
DeepLINK-T is available at \url{https://github.com/zuowx/DeepLINK-T}. Scripts for simulation studies and real data applications are also provided under the same GitHub repository.

\subsection*{Authors' contributions}
F.S., Y.F. and J.L. designed and supervised the study. W.Z. and Z.Z. implemented the method and conducted the simulation studies. W.Z. and Y.D. analyzed the real-world data sets. J.F. and Y.Y. advised on real-world data applications. W.Z. drafted the manuscript. All authors contributed to modify and finalize the manuscript. All authors read and approved the final version of the manuscript.

\end{document}


\title{Supplementary material to ``DeepLINK-T: deep learning inference for time series data using knockoffs and LSTM"}
\maketitle

\begin{center}	
Wenxuan Zuo, Zifan Zhu, Yuxuan Du, Yi-Chun Yeh, Jed A. Fuhrman, Jinchi Lv, Yingying Fan and Fengzhu Sun
\end{center}

\bigskip
\appendix

\section{The impacts of hyperparameters and model misspecification on DeepLINK-T} \label{new.secA}

\begin{figure}[t]%
\centering
\includegraphics[width=\textwidth]{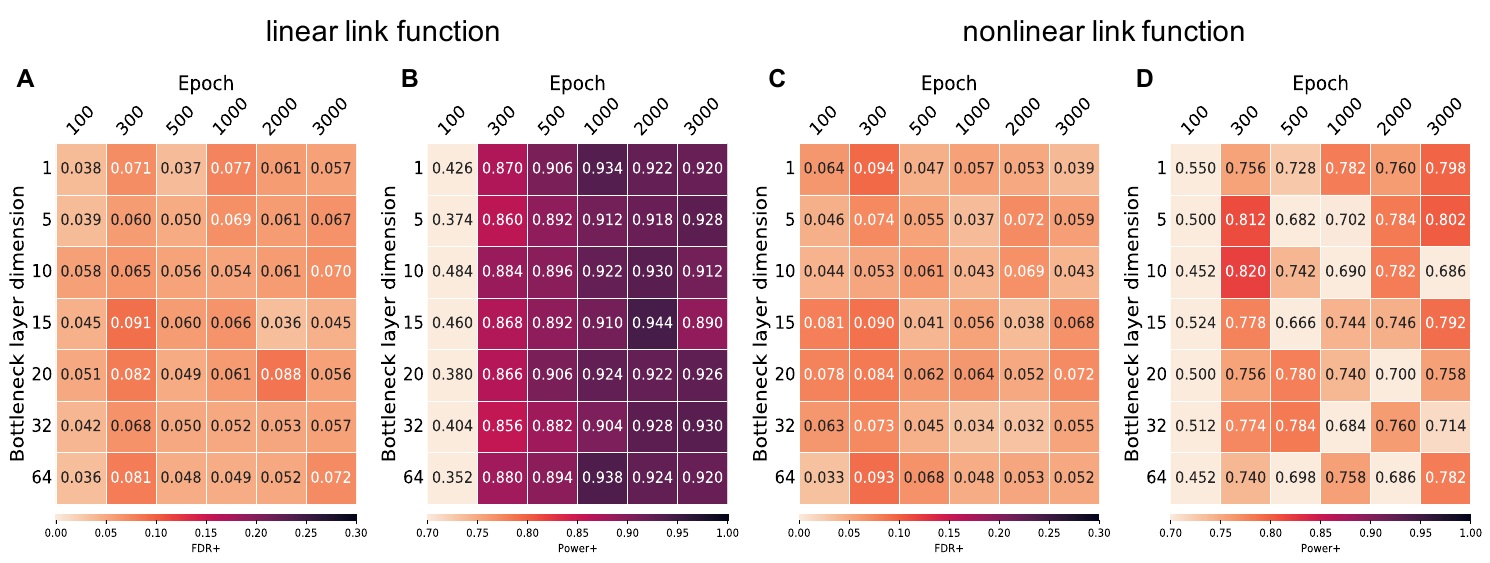}
\caption{The impacts of bottleneck dimensionality and training epochs on DeepLINK-T using the \textbf{linear} factor model. \textbf{Latent confouders} are added to make the regression model misspecified. A and B are FDR and power under the setting of linear link function. C and D are FDR and power under the setting of nonlinear link function. The number of training epochs of both LSTM autoencoder and LSTM prediction network is specified on the x-axis. The bottleneck dimensionality of the LSTM autoencoder is specified on the y-axis. The pre-specified FDR level is $q=0.2$.}\label{fig:sim_heatmap_fm_linear_confounder}
\end{figure}

\begin{figure}[t]%
\centering
\includegraphics[width=\textwidth]{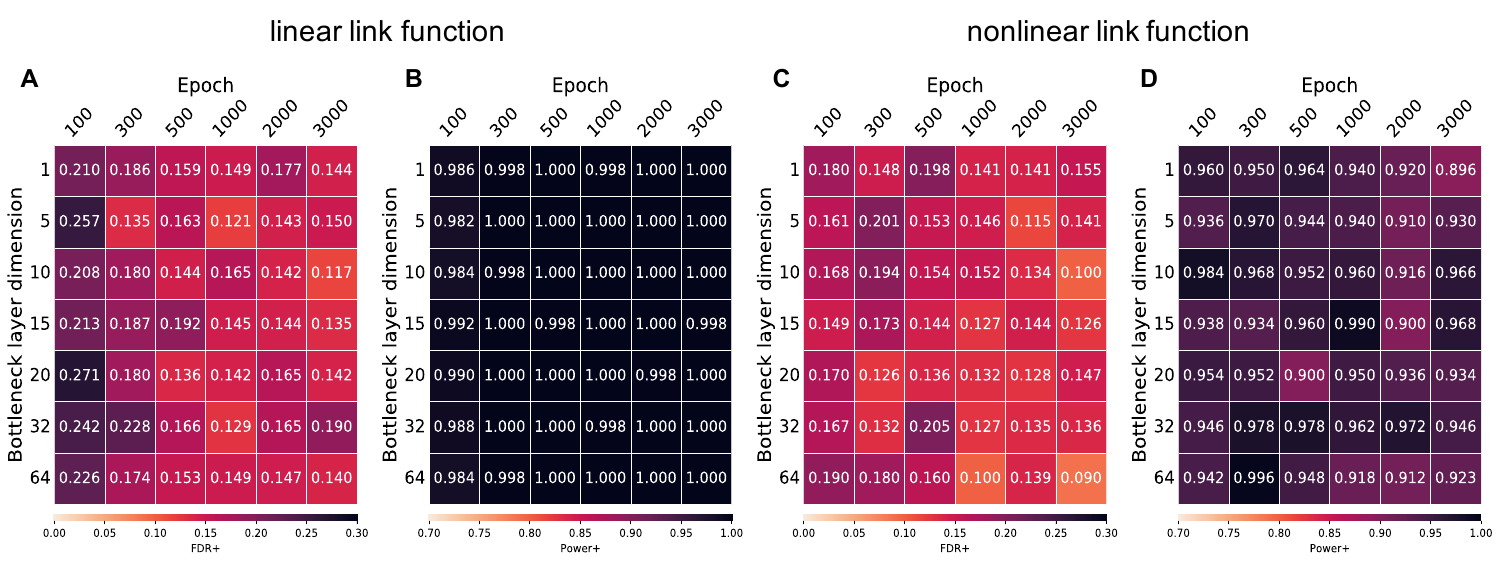}
\caption{The impacts of bottleneck dimensionality and training epochs on DeepLINK-T using the \textbf{logistic} factor model. \textbf{Latent confouders} are added to make the regression model misspecified. A and B are FDR and power under the setting of linear link function. C and D are FDR and power under the setting of nonlinear link function. The number of training epochs of both LSTM autoencoder and LSTM prediction network is specified on the x-axis. The bottleneck dimensionality of the LSTM autoencoder is specified on the y-axis. The pre-specified FDR level is $q=0.2$.}\label{fig:sim_heatmap_fm_nonlinear_confounder}
\end{figure}

To evaluate the ability of DeepLINK-T to deal with model misspecification, we introduced latent confounders into the linear or nonlinear regression model utilized in our simulation studies. Figures \ref{fig:sim_heatmap_fm_linear_confounder} and \ref{fig:sim_heatmap_fm_nonlinear_confounder} show the results using the linear and logistic factor models, respectively. Subfigures A and B present FDR+ and power+ with the linear link function, while subfigures C and D show FDR+ and power+ using the nonlinear link function. In each subfigure, the x-axis denotes the number of training epochs for both networks in DeepLINK-T and the y-axis indicates the bottleneck layer dimensionality in the LSTM autoencoder. Each cell in the heatmap reflects the mean value of FDR+ or power+ based on 50 runs of DeepLINK-T. After introducing the latent confounders, an increased number of epochs is required for the model to converge during the training process. Once the number of training epochs increases to 1000 or larger, DeepLINK-T demonstrates the ability to achieve high power while effectively controlling the FDR, even in the presence of model misspecification.

\section{Application to longitudinal gut microbiome data set of early infants} \label{new.secB}

\begin{figure}[t]%
\centering
\includegraphics[width=0.9\textwidth]{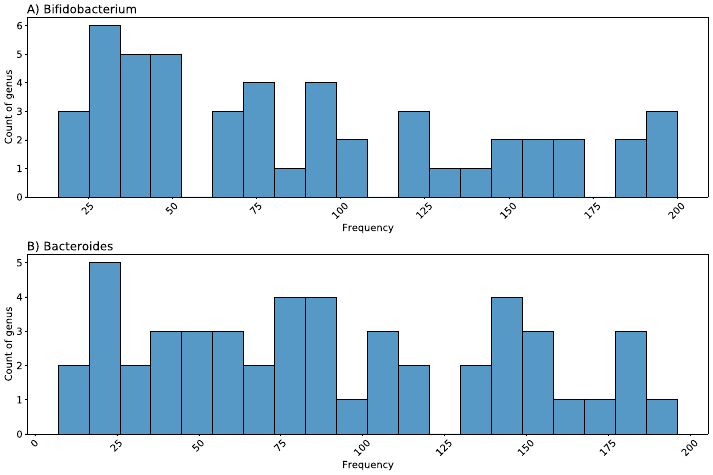}
\caption{The distribution of selection frequencies on the gut microbiome data set of early infants over 200 runs.}
\label{fig:ex1_histogram}
\end{figure}

\begin{figure}[t]%
\centering
\includegraphics[width=\textwidth]{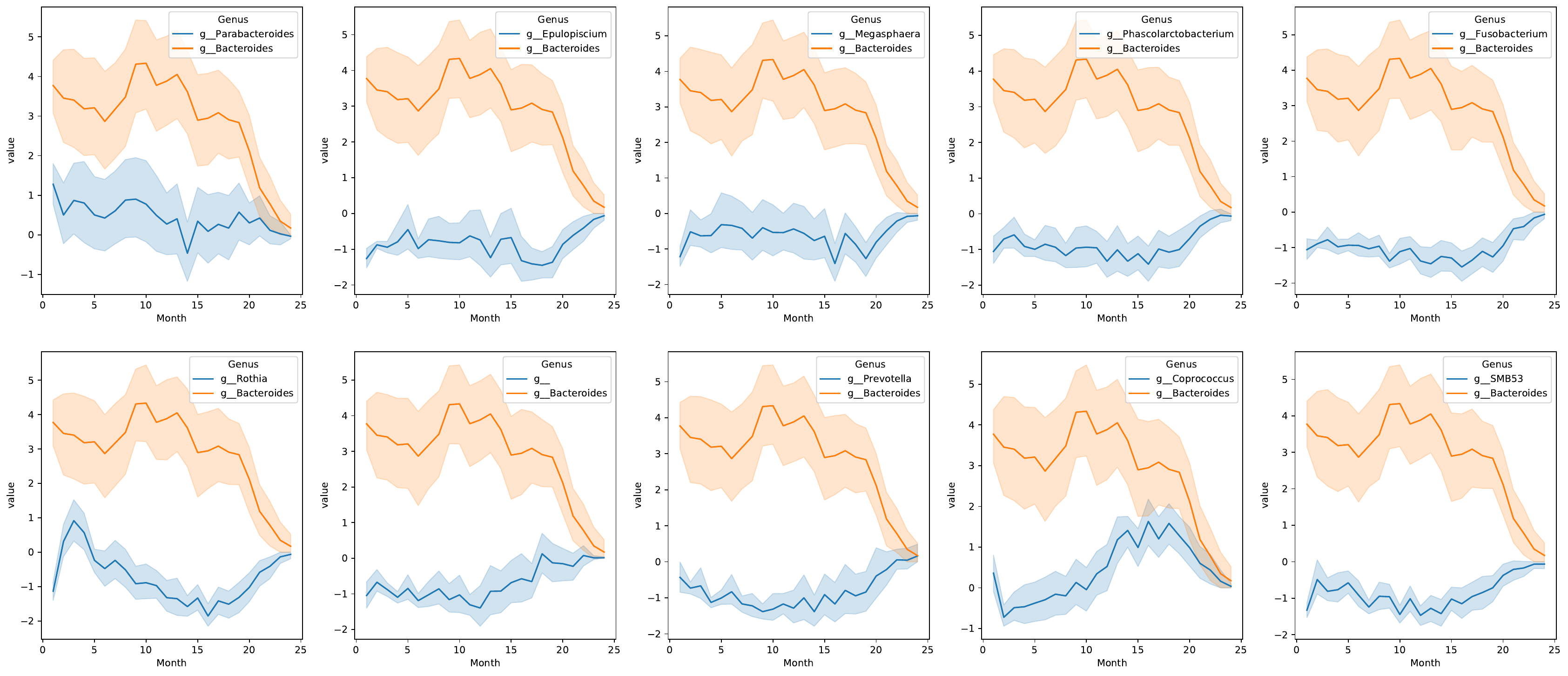}
\caption{The abundance variation of \textit{Bacteroides} and its top 10 selected genera.}
\label{fig:ex1_sclr_bac_ts_plot}
\end{figure}

\begin{figure}[t]%
\centering
\includegraphics[width=\textwidth]{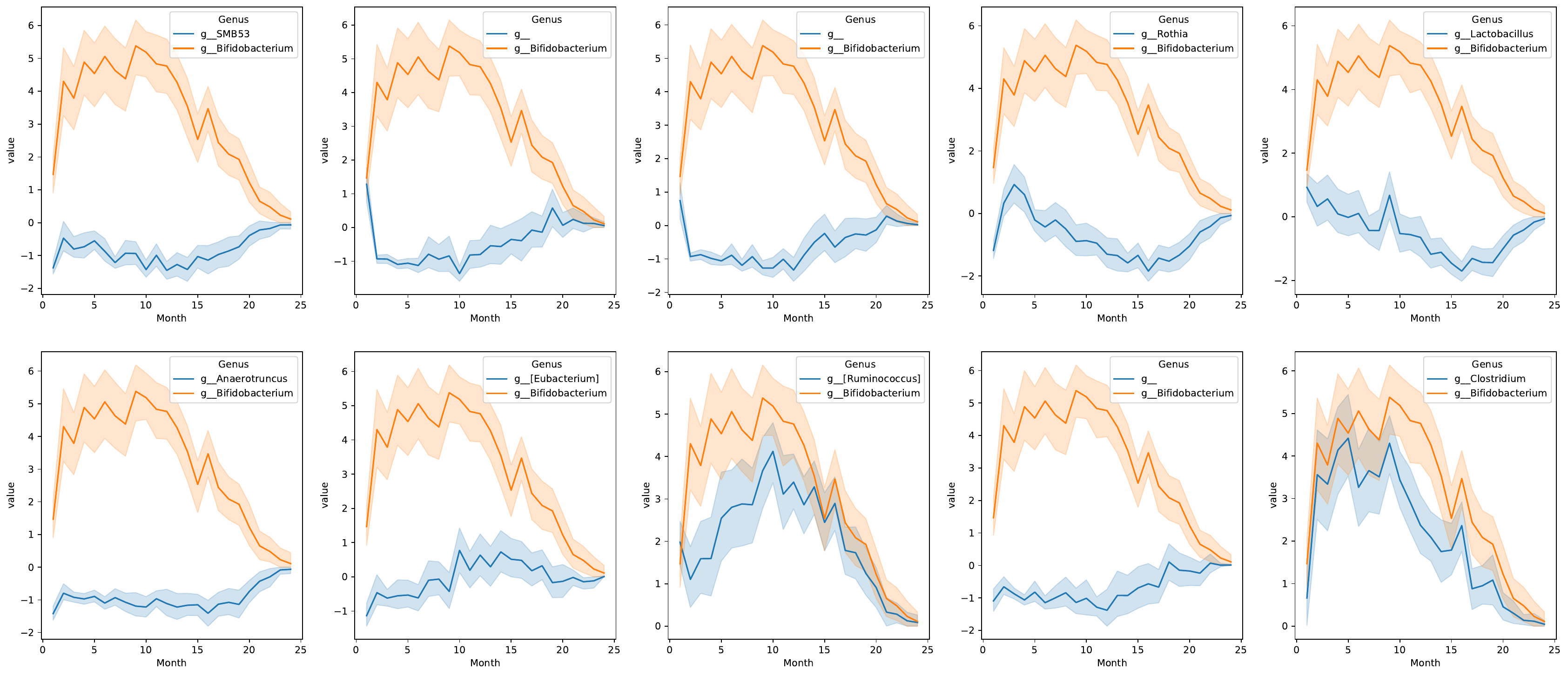}
\caption{The abundance variation of \textit{Bifidobacterium} and its top 10 selected genera.}
\label{fig:ex1_sclr_bif_ts_plot}
\end{figure}

To evaluate the performance of DeepLINK-T on real-world application, we first applied it to a longitudinal gut microbiome data set of early infants. We aimed to identify important bacterial genera associated with the
abundance levels of \textit{Bacteroides} and \textit{Bifidobacterium}. For each response genus, DeepLINK-T was fit on the data set 200 times. The histograms of selection frequencies for associated genera are shown in Figure \ref{fig:ex1_histogram}, which showcase that DeepLINK-T is able to consistently select significant bacterial genera associated with the response. The abundance variation of the top 10 selection genera along with the response is shown in Figures \ref{fig:ex1_sclr_bac_ts_plot} and \ref{fig:ex1_sclr_bif_ts_plot}. Most selected bacterial genera share similar or opposite variation patterns with their response genus.




\clearpage
\section{Application to marine metagenomic time series data set} \label{new.secC}

\begin{figure}[t]%
\centering
\includegraphics[width=0.9\textwidth]{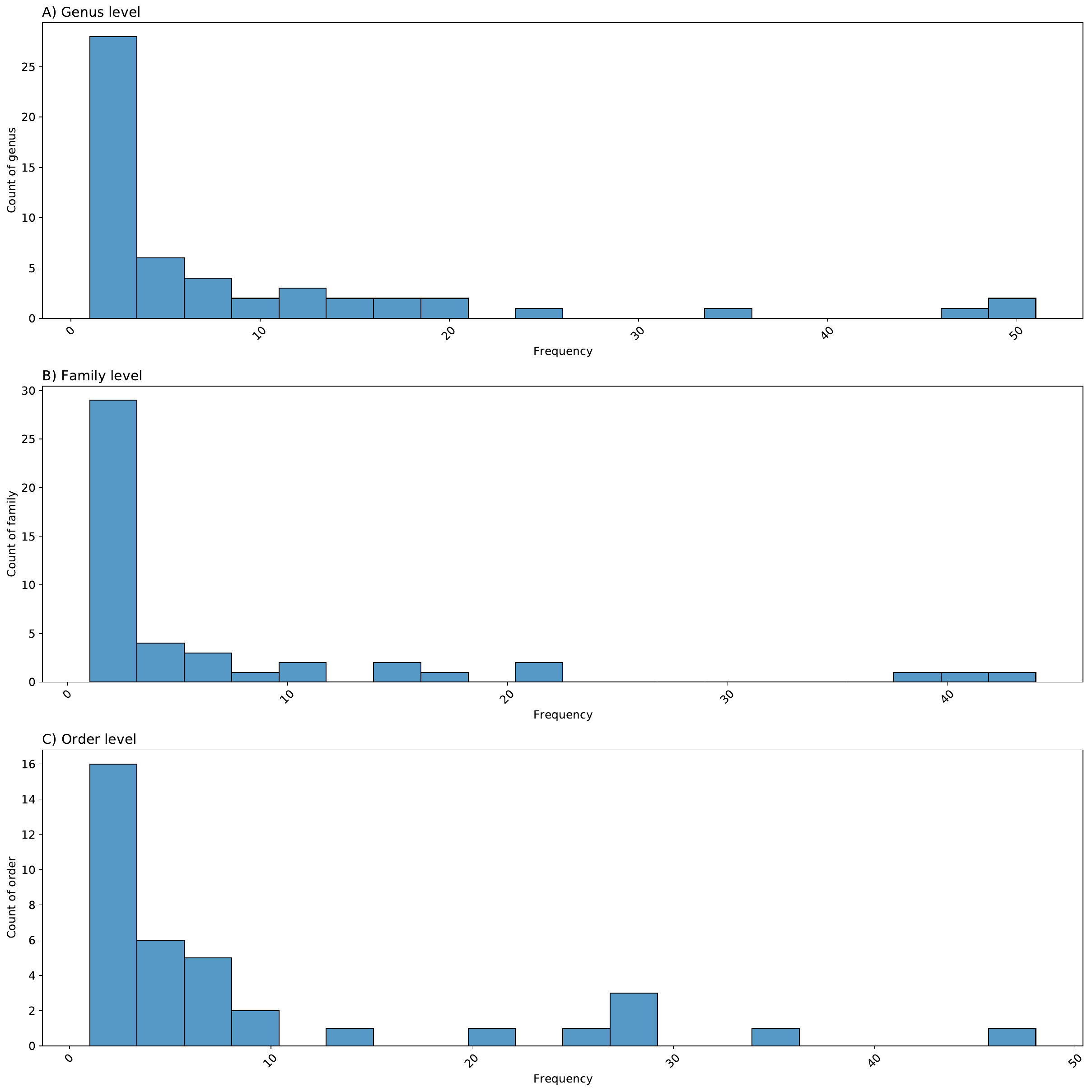}
\caption{The distribution of selection frequencies for response chlorophyll-a concentration on the SPOT data set over 200 runs.}
\label{fig:ex2_histogram}
\end{figure}

\begin{figure}[t]%
\centering
\includegraphics[width=\textwidth]{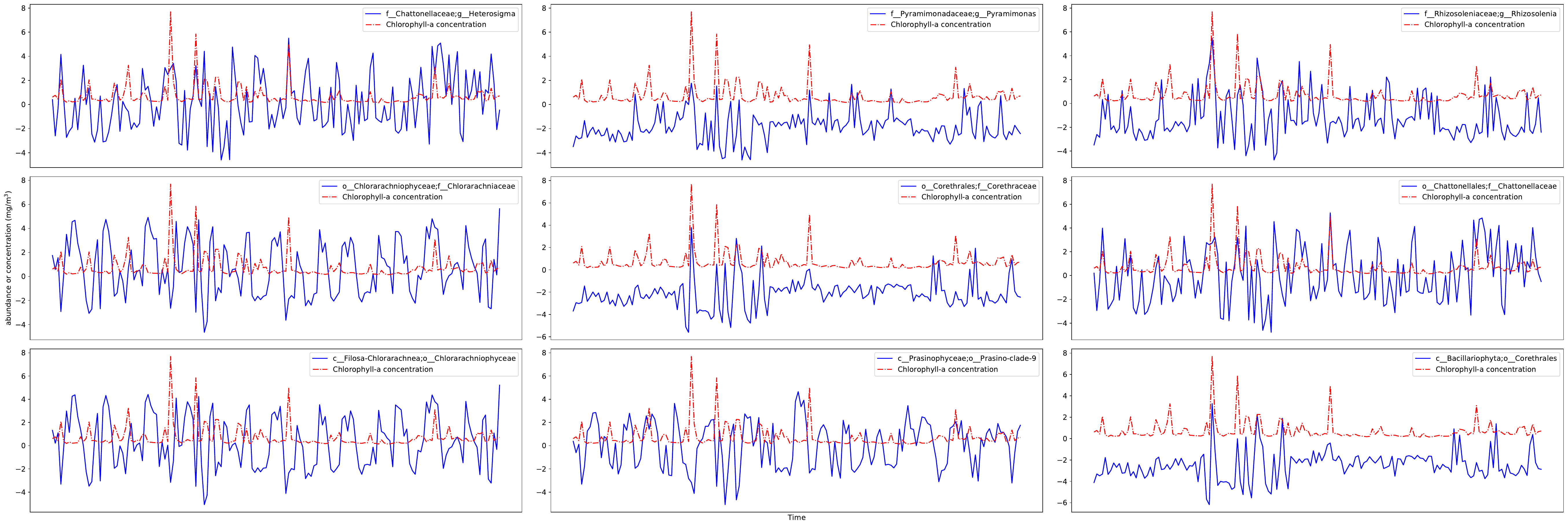}
\caption{The variation of chlorophyll-a concentration and the abundance of its top 3 selected taxa at the genus, family, and order levels.}
\label{fig:ex2_ts_plot}
\end{figure}

Next, we applied DeepLINK-T to a marine metagenomic time series data set from the San Pedro Ocean, where exploring the association between the abundance of chloroplast and chlorophyll-a concentration is our primary interest. Similar to the first application, Figure \ref{fig:ex2_histogram} shows the selection frequencies of important taxa associated with the chlorophyll-a concentration at the genus, family, and order levels. Although the top selection frequencies are not comparable to those in the first application, Figure \ref{fig:ex2_ts_plot} delineates that the variation in the abundance of top selected taxa did align with the variation in chlorophyll-a concentration.

\begin{figure}[t]%
\centering
\includegraphics[width=0.9\textwidth]{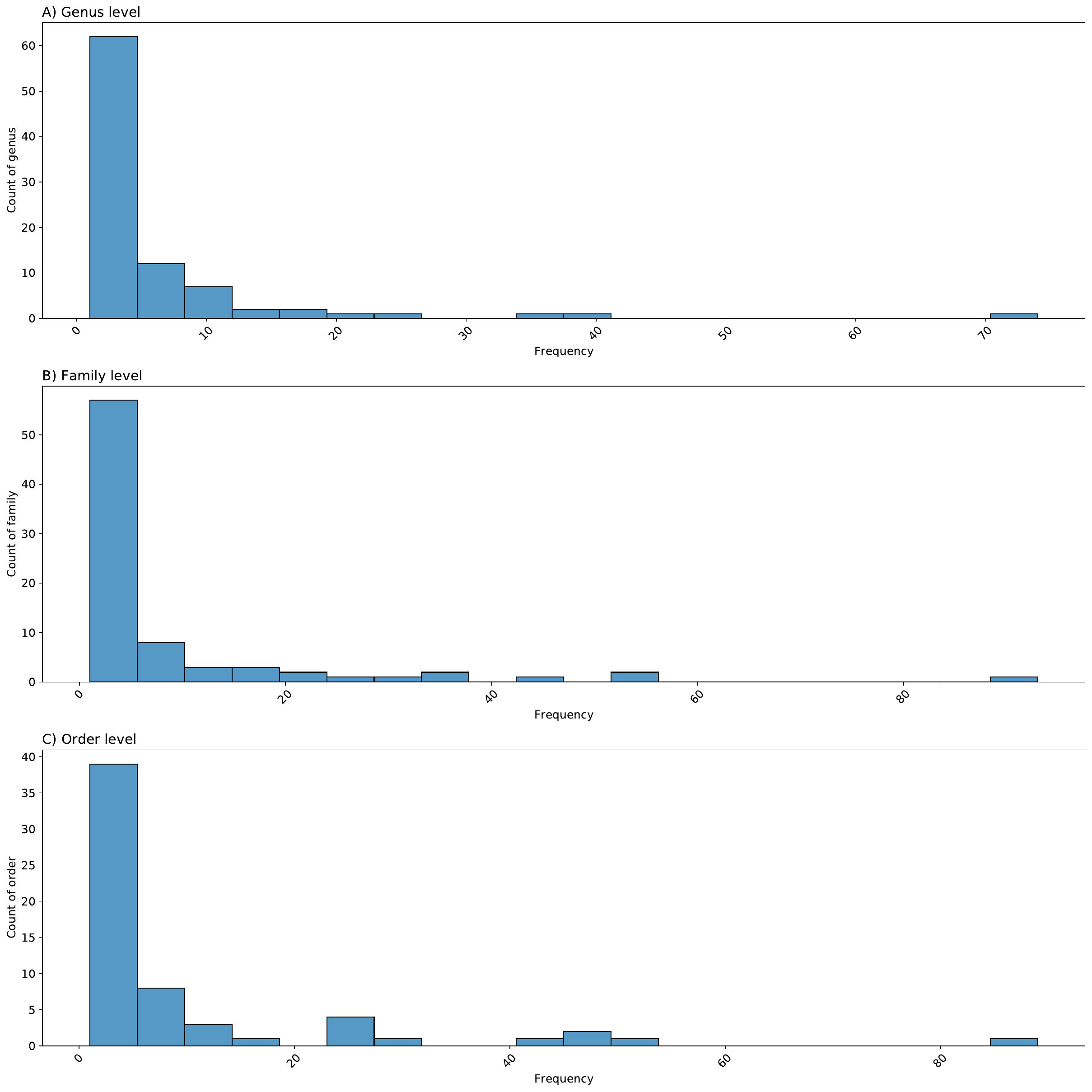}
\caption{The distribution of selection frequencies for response leucine incorporation on the SPOT data set over 200 runs.}
\label{fig:ex2_bpleu_histogram}
\end{figure}

We also leveraged the same dataset to investigate the relationship between the abundance of prokaryotes and leucine incorporation rate, a metric used to quantify the prokaryotic production. Similarly, Figure \ref{fig:ex2_bpleu_histogram} illustrates the selection frequencies of taxa associated with the leucine incorporation rate at the genus, family, and order levels. Notably, the top selection frequencies surpass those observed for chlorophyll-a concentration. Further insights are provided in Figure \ref{fig:ex2_ts_plot}, where the variation trend in the abundance of most top selected taxa closely aligned with fluctuations in the leucine incorporation rate.

\begin{table}[t]
\begin{tabular}{lc}
\toprule
                                         Genus &  Frequency \\
\midrule
 f\_\_Flavobacteriaceae;g\_\_Tenacibaculum &         74 \\
           f\_\_uncultured;g\_\_uncultured &         41 \\
  f\_\_Rhodobacteraceae;g\_\_Planktomarina &         36 \\
  f\_\_Flavobacteriaceae;g\_\_Polaribacter &         24 \\
   f\_\_Rhodobacteraceae;g\_\_Planktotalea &         22 \\
    f\_\_Rhodobacteraceae;g\_\_Amylibacter &         17 \\
         f\_\_Halieaceae;g\_\_Luminiphilus &         16 \\
                f\_\_Clade\_I;g\_\_Clade\_Ib &         13 \\
    f\_\_Flavobacteriaceae;g\_\_Aquibacter &         12 \\
f\_\_Flavobacteriaceae;g\_\_Psychroserpens &         11 \\
\midrule
                                                 Family &  Frequency \\
\midrule
       o\_\_Defluviicoccales;f\_\_uncultured &         93 \\
o\_\_Flavobacteriales;f\_\_Flavobacteriaceae &         55 \\
        o\_\_Cellvibrionales;f\_\_Halieaceae &         54 \\
   o\_\_Flavobacteriales;f\_\_Cryomorphaceae &         44 \\
   o\_\_Caulobacterales;f\_\_Hyphomonadaceae &         37 \\
  o\_\_Rhodobacterales;f\_\_Rhodobacteraceae &         34 \\
             o\_\_SAR11\_clade;f\_\_Clade\_III &         29 \\
          o\_\_Balneolales;f\_\_Balneolaceae &         26 \\
              o\_\_SAR11\_clade;f\_\_Clade\_IV &         23 \\
    o\_\_Cellvibrionales;f\_\_Porticoccaceae &         20 \\
\midrule
                                          Order &  Frequency \\
\midrule
                       c\_\_Alphaproteobacteria;o\_\_Defluviicoccales &         89 \\
                        c\_\_Alphaproteobacteria;o\_\_Caulobacterales &         51 \\
                        c\_\_Gammaproteobacteria;o\_\_Cellvibrionales &         47 \\
                        c\_\_Alphaproteobacteria;o\_\_Rhodobacterales &         46 \\
                                   c\_\_Rhodothermia;o\_\_Balneolales &         43 \\
                               c\_\_Bacteroidia;o\_\_Flavobacteriales &         30 \\
                      c\_\_Gammaproteobacteria;o\_\_Thiomicrospirales &         27 \\
                                              c\_\_BD7-11;o\_\_BD7-11 &         25 \\
                                  c\_\_Oligoflexia;o\_\_Oligoflexales &         25 \\
c\_\_Marinimicrobia\_(SAR406\_clade);o\_\_Marinimicrobia\_(SAR406\_clade) &         25 \\
\bottomrule
\end{tabular}
\caption{Selection frequencies of top 10 selected taxa for leucine incorporation at the genus, family, and order levels over 200 runs.}
\label{ex2_bpleu_table}
\end{table}

\begin{figure}[t]%
\centering
\includegraphics[width=\textwidth]{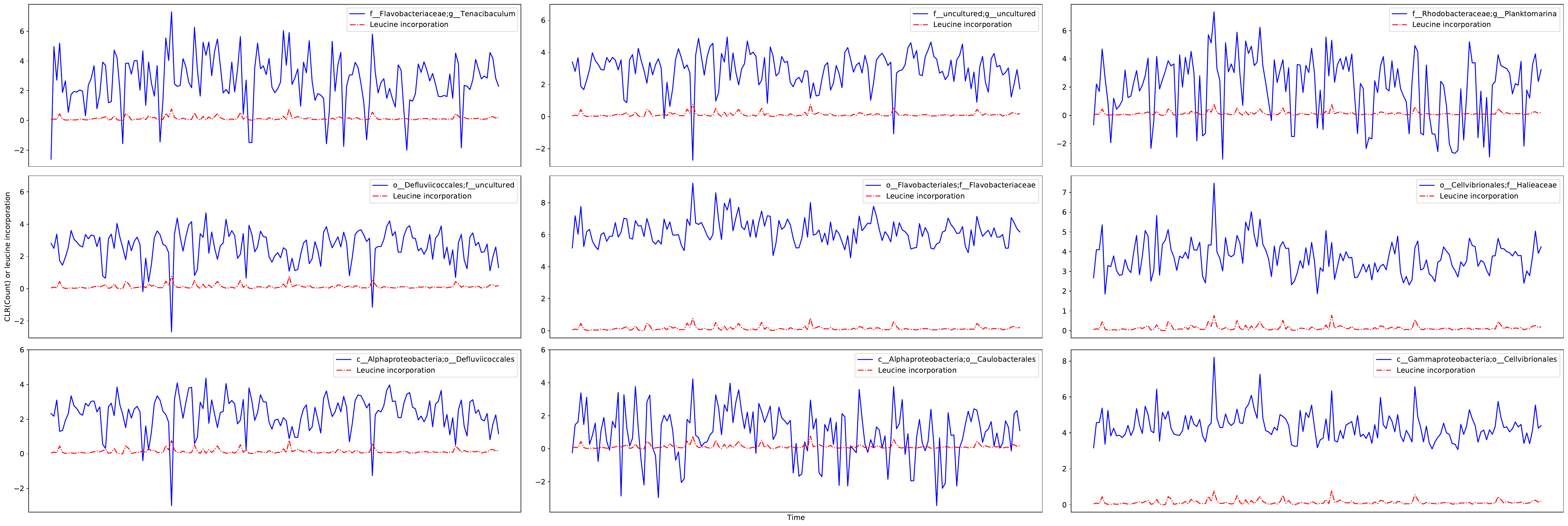}
\caption{The variation of leucine incorporation and the abundance of its top 3 selected taxa at the genus, family, and order levels.}
\label{fig:ex2_bpleu_ts_plot}
\end{figure}

\clearpage
\section{Application to dietary glycans treatment time series data set} \label{new.secD}

\begin{figure}[t]%
\centering
\includegraphics[width=0.9\textwidth]{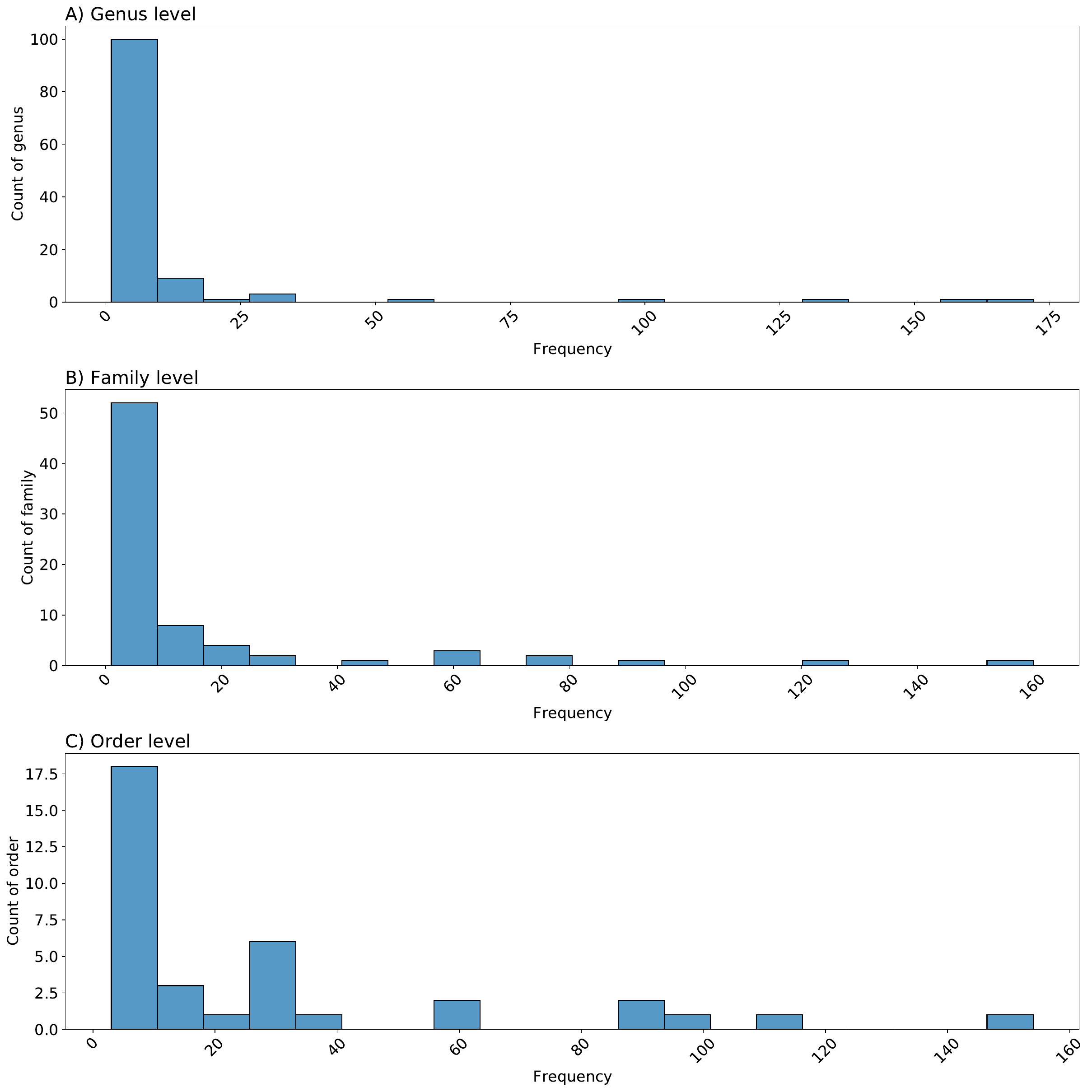}
\caption{The distribution of selection frequencies on the dietary glycans treatment time series data set over 200 runs.}
\label{fig:ex3_histogram}
\end{figure}

\begin{figure}[t]%
\centering
\includegraphics[width=\textwidth]{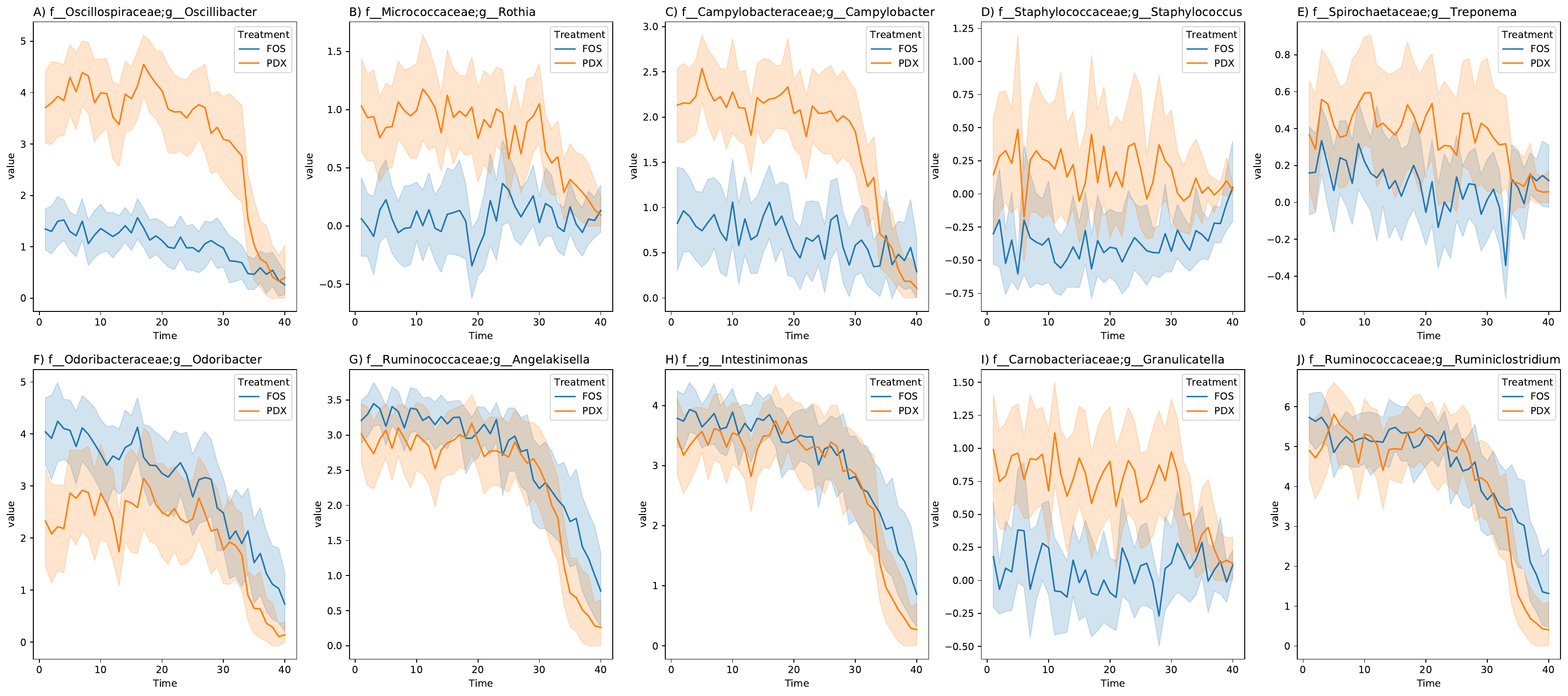}
\caption{The abundance variation of the top 10 selected genera between the FOS and PDX groups.}
\label{fig:ex3_genus_clr_ts_plot}
\end{figure}

\begin{figure}[t]%
\centering
\includegraphics[width=\textwidth]{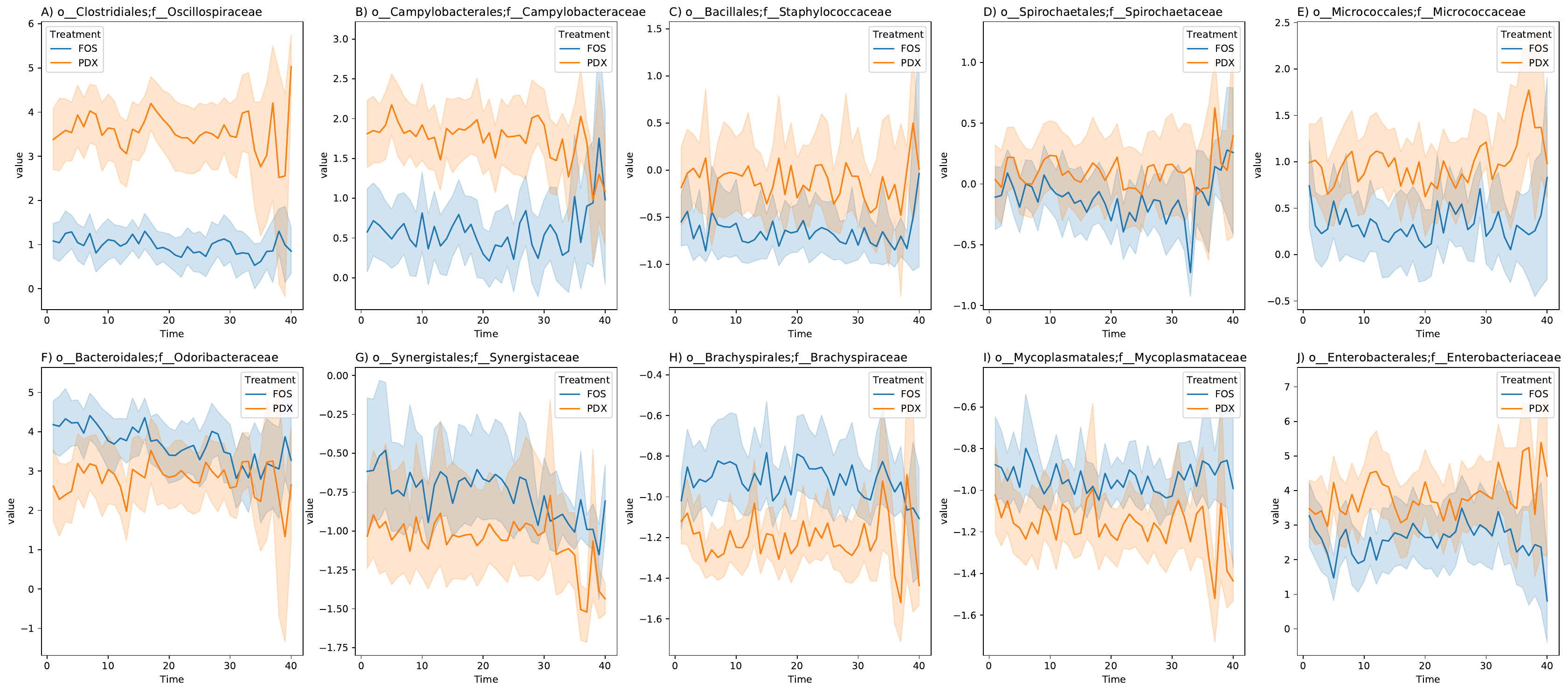}
\caption{The abundance variation of the top 10 selected families between the FOS and PDX groups.}
\label{fig:ex3_family_clr_ts_plot}
\end{figure}

\begin{figure}[t]%
\centering
\includegraphics[width=\textwidth]{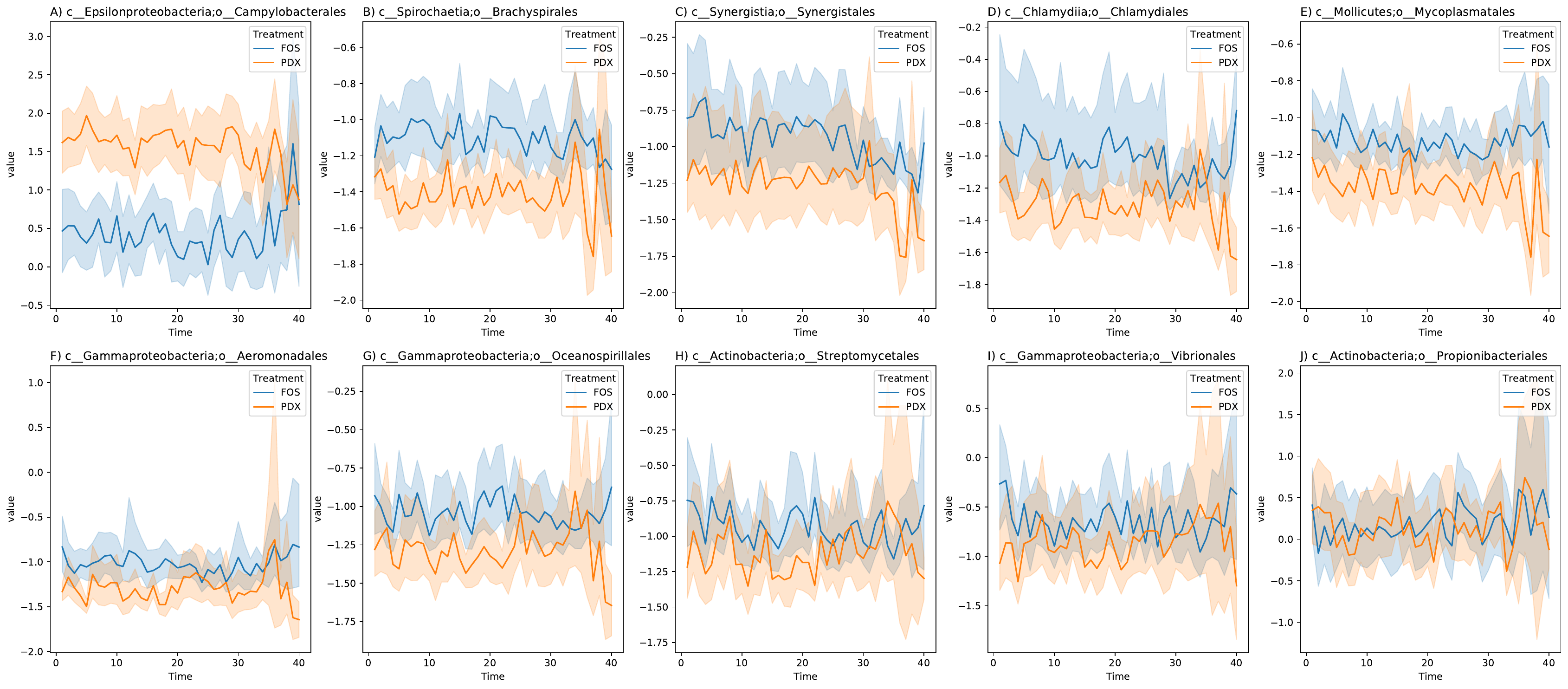}
\caption{The abundance variation of the top 10 selected orders between the FOS and PDX groups.}
\label{fig:ex3_order_clr_ts_plot}
\end{figure}

Finally, DeepLINK-T was applied to detect important bacterial taxa associated with two different dietary glycans treatments in a time series human gut metagenomic data set. Again, the application was implemented at the genus, family, and order levels. Figure \ref{fig:ex3_histogram} shows the selection frequencies of important taxa associated with fructooligosaccharides (FOS) and polydextrose (PDX) treatments at the three taxa levels. The results further demonstrate that DeepLINK-T can reliably select top features associated with the two treatments. The variation for the abundance level of top selected taxa categorized by FOS and PDX is shown in Figures \ref{fig:ex3_genus_clr_ts_plot}-\ref{fig:ex3_order_clr_ts_plot}, which additionally display the important bacterial taxa targeted by the two different treatments.

